\documentclass{article}

\usepackage[final]{arxiv}
\usepackage{graphicx}
\usepackage{epstopdf}
\usepackage{caption}

\usepackage[utf8]{inputenc} 
\usepackage[T1]{fontenc}    
\usepackage{hyperref}       
\usepackage{url}            
\usepackage{booktabs}       
\usepackage{amsfonts}       
\usepackage{nicefrac}       
\usepackage{microtype}      
\usepackage{setspace}       
\usepackage{amsmath}
\usepackage{amssymb}
\usepackage{adjustbox}
\usepackage{tablefootnote}
\usepackage[bottom]{footmisc}

\usepackage{algorithmic}
\usepackage{subcaption}
\usepackage[ruled]{algorithm2e}

\usepackage{natbib}
\title{Study Features via Exploring Distribution Structure}

\author{
Chunxu CAO \And Qiang ZHANG
}

\begin{document}

\begin{spacing}{2.0}

\maketitle

\begin{abstract}
In this paper, we present a novel framework for data redundancy measurement based on probabilistic modeling of datasets, and a new criterion for redundancy detection that is resilient to noise. We also develop new methods for data redundancy reduction using both deterministic and stochastic optimization techniques. Our framework is flexible and can handle different types of features, and our experiments on benchmark datasets demonstrate the effectiveness of our methods. We provide a new perspective on feature selection, and propose effective and robust approaches for both supervised and unsupervised learning problems.
\end{abstract}

\section{Introduction}

High-dimensional data, prevalent in many domains, offer rich and diverse information. Yet, they also present difficulties for machine learning and knowledge discovery tasks. These difficulties stem from the curse of dimensionality \cite{bellman1957dynamic}, overfitting, and the high storage and computation demands of high-dimensional data.

High-dimensional datasets pose several challenges for machine learning and knowledge discovery tasks. Feature selection is a common strategy to overcome these challenges by selecting relevant features from the original feature set and discarding noise and irrelevant features to reduce the dimensionality of the feature space \cite{guyon2003introduction}. The benefits of feature selection include data visualization \cite{yang1999data}, overfitting prevention, and storage and computational cost reduction. The field of feature selection has advanced rapidly in the past decades, and many algorithms have been developed and applied in various domains \cite{guyon2003introduction, cai2018feature, li2018feature}. These algorithms select an optimal feature subset from the original feature set using a certain goodness measure. Feature selection methods can be divided into two types: model-based and model-free. Model-based methods use specific learning models to evaluate features, and measure the goodness of feature sets based on the model performance or parameters. Model-free methods use data characteristics to evaluate features according to certain criteria. Many models and criteria have been used as feature evaluators \cite{cai2018feature}. These evaluators effectively measure the feature utility, and many provide interpretable and theoretically sound explanations.

Most feature selection methods can effectively measure the goodness or utility of features, but few can detect redundancy among them. This prevents them from producing an optimal minimal feature subset \cite{yu2004efficient}. Moreover, many methods ignore feature redundancy, which can degrade performance in subsequent analyses and limit knowledge discovery applications. This problem stems from the lack of consideration of feature interactions in many feature selection approaches. For instance, most similarity-based methods have difficulty dealing with feature redundancy \cite{zhao2013similarity, li2018feature}. These methods evaluate features based on their preservation of similarity across data. A unified framework can include many well-known similarity-based methods \cite{duda2000pattern, kira1992practical, he2005laplacian, zhao2007spectral, nie2008trace}, as shown in \cite{zhao2013similarity}.

Feature selection methods that fit within this framework assess the merit of a feature set by summing up the utility of each individual feature in the set. In other words, they evaluate features individually, which prevents them from effectively addressing feature redundancy.
Feature selection methods that fit within this framework evaluate the value of a feature set by summing up the utility of each individual feature in the set. In other words, they evaluate features individually, which prevents them from effectively addressing feature redundancy.

Indeed, some feature selection methods do take into account the interactions among features. \cite{yu2004efficient} suggested that the relevance and redundancy of features should be analyzed to develop an efficient feature selection algorithm. They also categorized the input features into four classes. Since then, a significant amount of work has been done focusing on redundancy has emerged.
These methods can be divided into different categories based on how they detect redundancy: model-based and model-free methods. Model-based methods, which use learning models to capture the complex interactions among features, are quite popular. They evaluate the feature set using a specific learning model, and quantify the usefulness of the feature set according to the performance or parameters of the models \cite{kohavi1997wrappers, yamada2014highdimensional, lemhadri2021lassonet}. However, these methods often face a trade-off between detection ability and efficiency. The more expressive the learning model is, the more powerful it is at detecting interactions. Another major drawback of model-based methods is their lack of interpretability, especially when it comes to analyzing feature redundancy explicitly. In addition, the generalization capacity of the feature set selected by model-based feature selection methods is strongly influenced by the model.

On the other hand, model-free feature selection methods mainly include statistical correlation and mutual information based methods \cite{yu2004efficient, hanchuanpeng2005feature, tuv2009feature, xu2017semisupervised, xu2022general, wollstadt2023rigorous}. These methods, based on statistical correlation and information theory tools, provide a more explicit and interpretable description of the relationship between features than model-based methods do. Statistical correlation-based methods evaluate the relationship between features and labels, as well as the interaction between features, by using specific statistical correlation measures \cite{hall1999feature, yu2003feature, yu2004efficient, song2012feature}. Information theoretic methods quantify the informativeness of features using tools originating from information theory, such as mutual information \cite{lewis1992feature, battiti1994using, yang1999data, fleuret2004fast, hanchuanpeng2005feature, meyer2006use, meyer2008informationtheoretic, estevez2009normalized, brown2012conditional}. Both statistical correlation and mutual information based criteria can be seen as measures of correlation between variables. Using these correlation measures, we can quantitatively analyze the correlation between models.

However, feature selection methods based on correlation measures have several disadvantages First, they may be limited in detecting complex modes of relationships. For example, Pearson correlation-based feature selection methods cannot measure the non-linear relationship among features, and thus cannot to detect redundancy. Kernel-enhanced correlation measures such as HSIC are more powerful tools, but their ability to detect nonlinear relationships and their efficiency highly depend on the choice of kernel \cite{song2012feature}. Second, even if the measure can detect complex relationships between variables, handling synergistic interactions among features remains challenging \cite{wollstadt2023rigorous}. Moreover, there is a dilemma that expressive correlation measures such as information theoretic criteria are prohibitively expensive to estimate, especially for datasets with high dimensionality. These limitations suggest the need for a new feature selection framework that allows explicit analysis of both feature relevance and redundancy.

In this paper, we introduce a new framework that utilizes similarity measures to detect redundancy in features. We propose to use metrics defined in probability space to reduce feature redundancy. We analyze the distinct behaviors that features exhibit across classes by measuring the distance between the corresponding class conditional empirical distributions to explore feature redundancy. In unsupervised learning settings, we utilize Gromov-Wasserstein distance( GWD) to measure the ability of features to retain information from the original dataset.

The main contributions of this paper are as follows:

\begin{itemize}
	\item We show that the relationship among features can be studied by focusing on the distances between empirical distributions.
	\item We propose two new frameworks for measuring redundancy in data based on a probabilistic modeling of data and illustrate that our similarity measure-based framework is flexible and extensible.
	\item We implement the proposed algorithms on popular benchmark datasets collected from various real-world domains and demonstrate superior performance.
\end{itemize}

\section{Related Works}

\subsection{Feature Selection Methods}
Feature selection methods can be broadly categorized into two classes based on the interaction between features and learning models in the feature selection process: model-based and model-free.

Model-based methods are popular for discovering relationships between variables in feature selection. The utility or goodness of features is quantified according to the performance or parameters of specific learning models. These methods evaluate features based on the training performance of specific learning models, and can be further divided into wrapper methods and embedded methods. Wrapper methods evaluate features based on the performance of a specific model trained on them, with the model playing an evaluator role in feature selection. Selected features are then fed into other models for further analysis. A typical method is SVM-RFE \cite{kohavi1997wrappers}. Embedded methods combine the model training process and the feature selection process, evaluating features and selecting important ones in a unified way. They are more effective than wrapper methods, but their ability to capture complex interactions relies on the expressiveness of the specific model. Many embedded methods utilize regularization learning techniques, such as HSIC-Lasso \cite{yamada2014highdimensional}, and LassoNet \cite{lemhadri2021lassonet}. Redundancy in such methods is usually presented as a small improvement or even reduction in model performance.

Model-free methods, also known as filters in literature, evaluate features according to the intrinsic information in features by means of certain criteria independent of learning models. Some methods aim to find features that best preserve the similarity structure in a dataset, such as Fisher Score \cite{duda2000pattern}, Laplace Score \cite{he2005laplacian}, SPEC \cite{zhao2007spectral}, and Trace Ratio \cite{nie2008trace}. These methods can be unified in a framework proposed by Zhao et al. \cite{zhao2013similarity}, who also pointed out that these similarity-based methods cannot handle interaction among features, thus failing to tackle redundancy. Some methods utilize information theoretic tools to construct criteria for evaluating features. A typical method is MIM (a.k.a. Information Gain) \cite{lewis1992feature}, which uses information theoretic tools to measure the correlation between features and labels and chooses the strongest ones. Information tools can also be used to detect redundancy in features, with some methods considering it via a redundancy term, such as MIFS \cite{battiti1994using}, MRMR \cite{hanchuanpeng2005feature} and CMIM \cite{fleuret2004fast}. Most of these information theoretic methods can be unified in a conditional likelihood maximization framework proposed by Brown et al. \cite{brown2012conditional}. There are also some methods that utilize statistical correlation to evaluate the relevance and redundancy of features, such as Hall's method \cite{hall1999correlationbased}, which can be enhanced by kernel trick to meet the demand for tackling non-linear relationship between features \cite{song2012feature}. Redundancy in model-free methods is usually quantifi
Model-free methods, also known as filters in the literature, evaluate features according to the intrinsic information in the features by means of certain criteria independent of learning models. Some methods aim at finding features that best preserve the similarity structure in a dataset, such as Fisher Score \cite{duda2000pattern}, Laplace Score \cite{he2005laplacian}, SPEC \cite{zhao2007spectral}, and Trace Ratio \cite{nie2008trace}. These methods can be unified in a framework proposed by Zhao et al. \cite{zhao2013similarity}, who also pointed out that these similarity-based methods cannot handle interaction among features, and thus failing to tackle redundancy. Some methods utilize information theoretic tools to construct criteria for evaluating features. A typical method is MIM (a.k.a. Information Gain) \cite{lewis1992feature}, which uses information theoretic tools to measure the correlation between features and labels and selects the strongest ones. Information tools can also be used to detect redundancy in features, with some methods accounting for it via a redundancy term, such as MIFS \cite{battiti1994using}, MRMR \cite{hanchuanpeng2005feature} and CMIM \cite{fleuret2004fast}. Most of these information theoretic methods can be unified in a conditional likelihood maximization framework proposed by Brown et al. \cite{brown2012conditional}. There are also some methods that utilize statistical correlation to evaluate the relevance and redundancy of features, such as Hall's method \cite{hall1999correlationbased}, which can be extended by kernel tricks to meet the need to tackle non-linear relationships between features \cite{song2012feature}. Redundancy in model-free methods is usually quantified as shared information or little change in correlation.
ed as shared information or little change of correlation.

\subsection{Literature for Redundancy Reduction}

\subsubsection{Measuring the Redundancy}
Using a learning model to measure the goodness of features is an intuitive and effective idea. Clearly, a feature is redundant if the performance of a learning model trained on selected features does not change significantly. However, model-based methods cannot explicitly handle redundancy in features along with their relevance \cite{yu2004efficient}.

Correlation measures are effective tools for detecting relationships in variables, with redundancy in features represented as correlation.  \cite{hall1999feature} utilized Pearson correlation and mutual information to quantify feature-label correlation and inter-feature correlation, and evaluated features based on estimated correlation.  \cite{yang1999data} used joint mutual information to tackle relevance and redundancy in a unified way.
\cite{yu2004efficient} formally defined relevance and redundancy, and introduced a correlation measure-based feature selection algorithm.
\cite{meyer2008informationtheoretic} introduced the use of Double Input Symmetrical Relevance (DISR) in quantifying redundancy and feature evaluation. 
\cite{estevez2009normalized} suggested normalizing mutual information by the minimum entropy of both features to eliminate the bias of mutual information.
\cite{brown2012conditional} proposed a unified framework that covers many popular information theoretic feature selection methods, and provided a general form that considers the relevance and redundancy of features.
\cite{zhao2013similarity} showed that many filter methods could be covered by a similarity-based framework and provided a correlation regularization-based method that could consider the interaction among features. \cite{yamada2014highdimensional} extended LASSO by HSIC to select features in high-dimensional datasets, detecting feature interaction by HSIC, a kernel-enhanced correlation measure.
\cite{wollstadt2023rigorous} pointed out that the weakness of mutual information based feature selection criteria in dealing with groups of variables is due to the inherent limitation of basic mutual information, and introduced the partial information decomposition (PID) framework to rigorously define redundancy.

\subsubsection{Strategies for Reducing the Redundancy}

Feature selection methods that consider interactions among features can also be divided into two categories based on how they deal with redundancy in features: two-stage methods and unified methods. The former first consider the relevance between features and labels to generate a set containing highly relevant features, and then reduce the redundancy in this set \cite{battiti1994using, yu2004efficient, chow2005estimating, hanchuanpeng2005feature, estevez2009normalized, brown2012conditional, poczos2012copulabased, yamada2014highdimensional, wollstadt2023rigorous}. The latter take relevance and redundancy into account simultaneously. Some do this by combining the relevance between features and labels and the redundancy in features into a single formula, while others leverage the inherent properties of specific distributions and models \cite{novovicova1996divergence, kohavi1997wrappers, yang1997comparative, tuv2009feature, zhao2013similarity}.

Two-stage methods can often be rewritten as follows, considering the interaction among features by an incremental search:
\begin{equation}
	Score(\mathcal{T}) = Relevance(\mathcal{T}) - Redundancy(\mathcal{T})
\end{equation}
This canonical form is intuitive and flexible. It can be compatible with various definitions and measures of relevance and redundancy.

Unified methods are implemented implicitly in the feature selection process. They are based on specific models, data consumption, and powerful theoretical tools that can simultaneously take relevance and redundancy into account. Model-based methods are evaluators of the utility of a feature set. They merge relevance and redundancy into a single evaluation metric: the performance or critical parameters of a model on a given feature set \cite{tuv2009feature}. An interesting example of unified methods based on data consumption is shown in \cite{novovicova1996divergence}, where the authors use a mixture of Gaussian distributions to implicitly detect and reduce feature redundancy. As for theoretical tools, in \cite{yang1999data}, the authors use joint mutual information to eliminate redundant input.

\section{Problem Formulation}
Reducing redundancy in features is a significant challenge, mainly due to the difficulty of defining and measuring redundancy.

Consider a data-generating model $P(\mathbf{x}, y)$ over a $d$-dimensional space, where $\mathbf{x} \in \mathcal{X} \subset R^d$ is the covariate and $y \in \mathcal{Y}={\{c_1,\dots,c_K\} }$ is the response, such as class labels. The goal of supervised learning is to find the best function $f^\star(\mathbf{x}) $ from the function class $\mathcal{F}$ to predict ${y}$. To measure the goodness of the prediction, the loss function $\mathcal{L}$ is used to quantify the difference between the prediction $f(\mathbf{x})$ and the ground truth $y$. We aim to find a function $f^\star(\mathbf{x}) $ that achieves the minimum risk $R_{exp}(f)$, which is the loss $\mathcal{L}$ in the expectation sense, among all the candidate functions in $\mathcal{F}$.

\begin{equation}
	f^{\star} =\underset{f \in \mathcal{F}}{\arg\min} R_{exp}(f) = \underset{f \in \mathcal{F}}{\arg\min} E[\mathcal{L}(f(\mathbf{x}), y)]
	\label{supervised_learning_problem}
\end{equation}

Feature selection aims to find the feature subset $\mathcal{T}^\star$ with the highest utility $U$ from the original feature set $\mathcal{S}$, which achieves the minimum risk $R_{exp}(f)$ among all the candidate feature subsets $\mathcal{T}$ with the required size $m$.

\begin{equation}
	\mathcal{T}^\star = \underset{ \mathcal{T} \subset \mathcal{S} ,f \in \mathcal{F}}{\arg\max} U(\mathcal{T}), \quad |\mathcal{T}|=m
	\label{feature_selection_problem}
\end{equation}

where $|\mathcal{T}|$ is the cardinality of the set $\mathcal{T}$. In supervised learning case, it's to find the feature subset $\mathcal{T}^\star$, on which we could find a best function $f^\star(\mathbf{x}_{\mathcal{T}}) $ from $\mathcal{F}$ to achieve the minimum risk $R_{exp}(f)$ in the expectation sense among all the candidate feature subsets.

\begin{equation}
	\begin{aligned}
		\mathcal{T}^\star & = \underset{ \mathcal{T} \subset \mathcal{S} ,f \in \mathcal{F}}{\arg\max} U(\mathcal{T}) \\ & = \underset{ \mathcal{T} \subset \mathcal{S} ,f \in \mathcal{F}}{\arg\min} R_{exp}( f(\mathbf{x}_{\mathcal{T}}) ) \\ & = \underset{ \mathcal{T} \subset \mathcal{S} ,f \in \mathcal{F}}{\arg\min} E[\mathcal{L}(f(\mathbf{x}_{\mathcal{T}}), y)], \quad |\mathcal{T}|=m
		\label{supvised_feature_selection_problem}
	\end{aligned}
\end{equation}

In practice, however, we often need to evaluate features without training a classifier. Therefore, we need to choose an appropriate quantization function $U(\cdot)$ and a selection strategy independent of learning models to find the optimal subset $\mathcal{T}^\star$.

Recent works have shown that the distance between class conditional empirical distributions bounds the performance of classifiers in certain function classes. \cite{nguyen2009surrogate} showed that for each loss function $L$ in a binary classification problem, there exists a corresponding $\phi$-divergence, the negative of which between different class conditional distributions is equal to the optimal $L$-risk. \cite{sriperumbudur2009integral} related Integral Probability Metrics (IPMs) to the optimal $L$-risk of a binary classification problem. A natural assumption is that the distance between class conditional probabilities still matters in multi-class settings. Therefore, the utility function $U$ could be set to the summarized differences between the class conditional distributions. Thus, equation (\ref{supvised_feature_selection_problem}) could be transformed as follows:
\begin{equation}
	\mathcal{T}^{\star} = \arg\max_{\mathcal{T}\subset S} \sum_{i}^{K}\sum_{j}^{K} {D} ( p^{\prime}(\mathbf{X}_{\mathcal{T}}|y=c_i) , p^{\prime}(\mathbf{X}_{\mathcal{T}}|y=c_j) |)
	\label{feature_expect_discrepancy_maximization}
\end{equation}
that the goal of feature selection is to select features possessing maximum of disparity, it also suggests that the distinguishing behavior of features across classes is the core.
This leads to our main results, exploring feature redundancy by studying disparity.

\begin{figure}[ht]
	\centering
	\begin{subfigure}[h]{0.49\textwidth}
		\includegraphics[width=\textwidth]{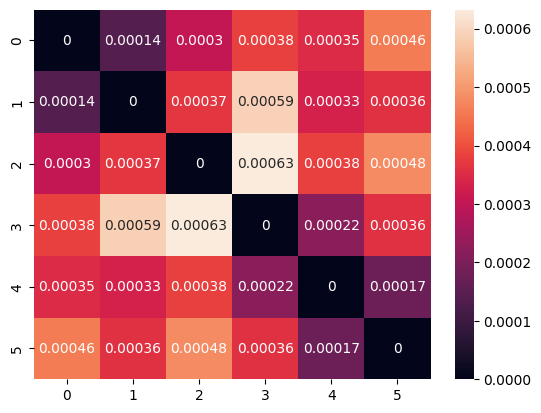}
		\caption{Normal noise.}
	\end{subfigure}
	\begin{subfigure}[h]{0.49\textwidth}
		\includegraphics[width=\textwidth]{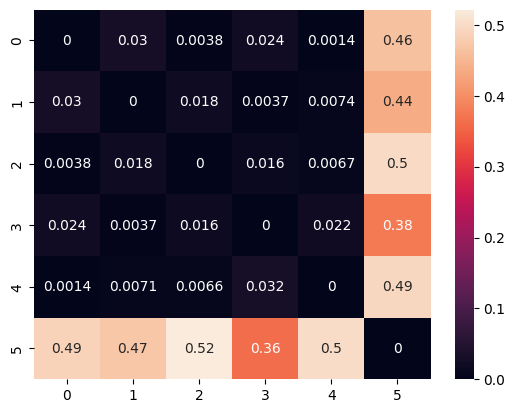}
		\caption{Relevant feature.}
	\end{subfigure}
	\caption{Distance matrices of relevant and irrelevant features. Relevant feature possesses much more significant disparity than random features.}
	\label{dist_matrix_relev_noise_1fs}
\end{figure}

\section{Methods}

\subsection{Similarity between distance matrices}
A straightforward approach to explore redundancy in features is to compare the distances across empirical distributions corresponding to different classes of the candidate feature set. This idea is based on previous analyses, which suggest that the distances between class conditional distributions are central to their utility in supervised classification problems.

Given a dataset $(\mathbf{X}, \mathbf{Y})$ sampled from the underlying generation distribution $P(\mathbf{x}, y), \mathbf{x} \in R^d$, the dataset matrix $\mathbf{X}$ consists of $n$ samples $\mathbf{x}_i, i = 1,\dots,n$ and each sample corresponds to a label $y_i \in {C_1,\dots,C_k}$. The distance between class conditional empirical distributions can be summarized in a distance matrix $\mathbf{D}\in R^{k\times k}$, where each element $\mathbf{D}_{i,j}$ is the distance between the conditional empirical distributions corresponding to class $i$ and $j$ respectively.

\begin{equation}
	\mathbf{D}_{ij}^\mathcal{T}=D( p^\prime(\mathbf{X}_{\mathcal{T}}|y=c_i), p^\prime(\mathbf{X}_{\mathcal{T}}|y=c_j))
\end{equation}

Considering the good theoretical justification of Integral Probability Metrics (IPMs) and the physical interpretation of optimal transport, we use the 1-Wasserstein distance as the distance measure, which is a special case of the $p$-Wasserstein distance.

\begin{equation}
	W_{p}(P ,Q)= (~ \underset{\gamma \in \Gamma(P ,Q)} {\inf}  \int _{\chi \times \chi}||x-y||^{p}d \gamma(x,y) ~)^{1/p}
	\label{p_wsd_definition}
\end{equation}

The Frobenius norm of the distance matrices corresponding to the feature set $\mathcal{T}$ could be used as a utility function to quantify its goodness in classification tasks.

\begin{equation}
	U(\mathcal{T}) = ||\mathbf{D}^\mathcal{T}||_{Frob}^2 = \sum_{i}\sum_{ j }(\mathbf{D}_{ij}^\mathcal{T})^2 = \mathbf{tr}\mathbf{D}^\mathcal{T} (\mathbf{D}^\mathcal{T})^T
	\label{multi_class_wsd1_imp}
\end{equation}

Furthermore, by considering the element-wise similarity between the distance matrices corresponding to features or feature sets, we can quantify the similarity of the utility of features or feature sets in classification tasks. Based on this, given a feature set, we can measure the redundancy among its internal features.

\subsection{Extension to Unsupervised Feature Selection}
In the case of unsupervised learning, the goal is to find the feature subset $\mathcal{T}^\star$ that preserves the most information from the original dataset. From a matrix approximation perspective, as many neural network based methods formulate feature selection in this form \cite{lemhadri2021lassonet, balin2019concrete, wu2021fractal}, the goal is to find a column index set $\mathcal{T}^\star$ to select a submatrix $\mathbf{X}_{\mathcal{T}}$ from the original data matrix $\mathbf{X}$ that can best reconstruct $\mathbf{X}$.

\begin{equation}
	\mathcal{T}^\star = \underset{\mathcal{T}}{\arg\min} L(\theta,\mathcal{T}) = || f_{\theta} \mathbf{X}_{\mathcal{T}} - \mathbf{X}||_F^2
\end{equation}

where $f_{\theta}$ is a parametric model that can reconstruct the matrix, and $\theta$ is its parameter. From a probabilistic perspective, if we focus on approximating the original high-dimensional generation distribution with a low-dimensional distribution that inherits some of the original coordinates, the problem is to minimize the distance between them.

\begin{equation}
	\mathcal{T}^\star = \underset{\mathcal{T}}{\arg\min} D(P,Q)
\end{equation}

where $Q$ is the low-dimensional approximation of $P$. Finding such an optimal set could be NP-hard, but we can approximate the optimal set using heuristic strategies.

We can quantify the redundancy $R$ of a given feature $\mathbf{f}_t$ for feature set $\mathcal{T}$ as follows:

\begin{equation}
	R(\mathbf{f}_t) = \frac{1}{D(P_{\mathcal{T} \setminus \mathbf{f}_t },  P_{\mathcal{T}}) }
\end{equation}

In practice, we estimate the dissimilarity between distributions by a given dataset $\mathbf{X}$ is sampled from the higher dimensional distribution:

\begin{equation}
	\hat{R}(\mathbf{f}_t) = \frac{1}{D(\mathbf{X}_{\mathcal{T} \setminus \mathbf{f}_t },  \mathbf{X}_{\mathcal{T}}) }
\end{equation}

It's intuitive to say that such a feature is redundant if the distance between the class conditional distributions corresponding to feature sets that contain or do not contain that feature is small.
The challenge in this approach is to measure the distance between probability measures that exist in spaces of different dimensions. A typical measure to address this is the Gromov-Wasserstein distance \cite{memoli2011gromov}, which is a dissimilarity measure between metric measure spaces. It takes geometric information into account, similar to the $p$-Wasserstein distance, while also being able to compare distributions that exist in different metric spaces.

Given two metric measure spaces $(\mathcal{X}, d_{\mathcal{X}}, \mu_{\mathcal{X}} )$ and  $(\mathcal{Y}, d_{\mathcal{Y}}, \mu_{\mathcal{Y}} )$, the $(p,q)$-Gromov-Wasserstein distance is defined as follows:

\begin{equation}
	\mathrm{D}_{p, q}(\mu, \nu):=\inf _{\pi \in \Pi(\mu, \nu)}\left(\int_{\mathcal{X} \times \mathcal{Y}} \int_{\mathcal{X} \times \mathcal{Y}}\left|d_{\mathcal{X}}\left(x, x^{\prime}\right)^q-d_{\mathcal{Y}}\left(y, y^{\prime}\right)^q\right|^p \mathrm{~d} \pi \otimes \pi\left(x, y, x, y^{\prime}\right)\right)^{\frac{1}{p}}
	\label{gw_distance}
\end{equation}

where $\Pi(\mu, \nu)$ is the set of all couplings between two Borel probability measures $\mu$ and $\nu$.

This approach is flexible: the dissimilarity measure could be replaced by other suitable measures, and it can be directly extended to supervised learning settings. Comparing class conditional empirical distributions is also feasible.

\section{Experiments}
In this section, we conduct experiments\footnote{The github URL will be provided shortly.} to evaluate the effectiveness of the proposed methods. We apply them to real-world datasets and benchmark them against state-of-the-art feature selection methods. We show that the proposed methods achieve competitive performance with the existing methods.

\begin{figure}[!ht]
	\centering
	\begin{subfigure}[t]{0.32\textwidth}
		\includegraphics[width=\textwidth]{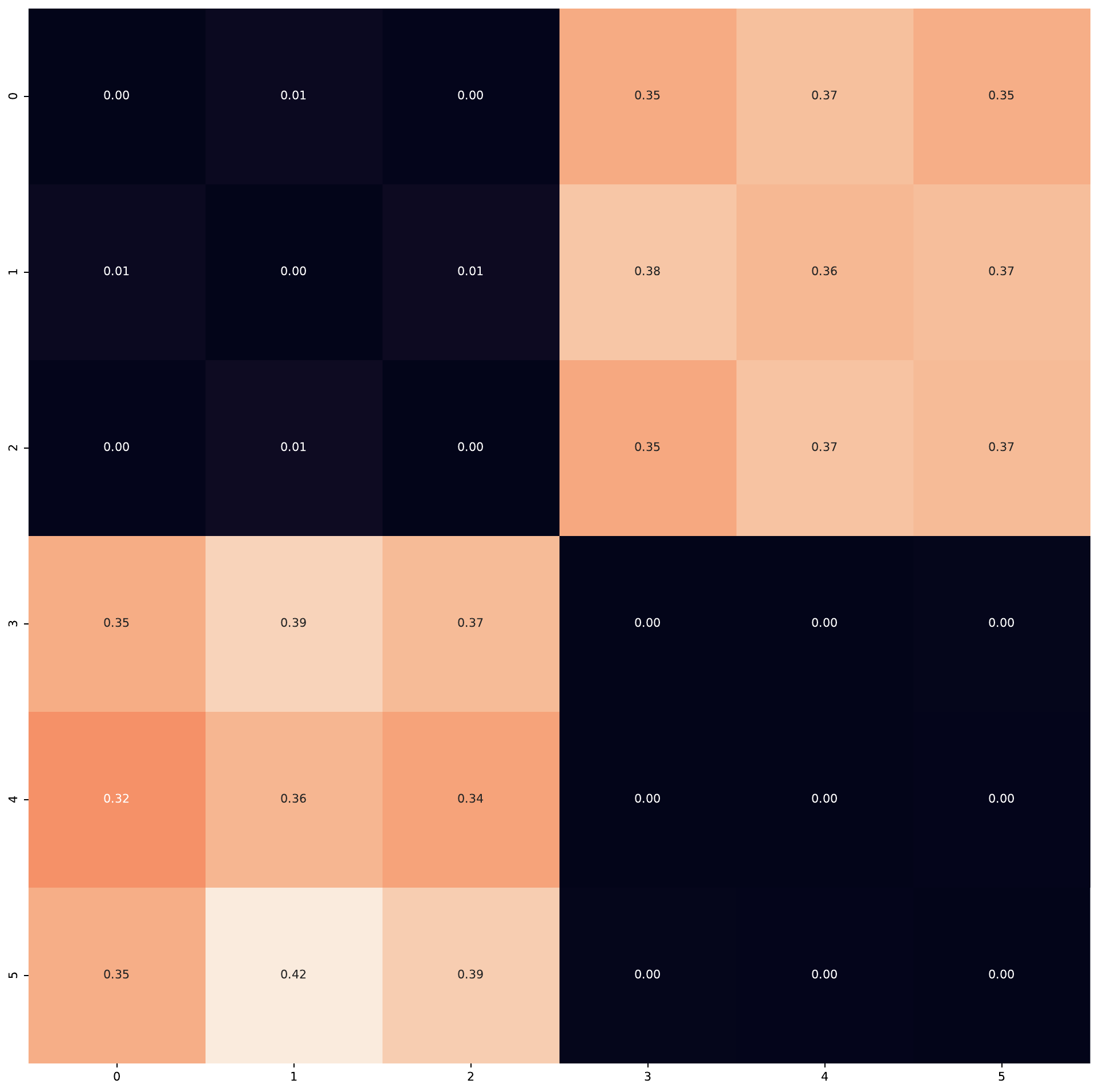}
		\caption{Top feature in Activity.}
	\end{subfigure}
	\begin{subfigure}[t]{0.32\textwidth}
		\includegraphics[width=\textwidth]{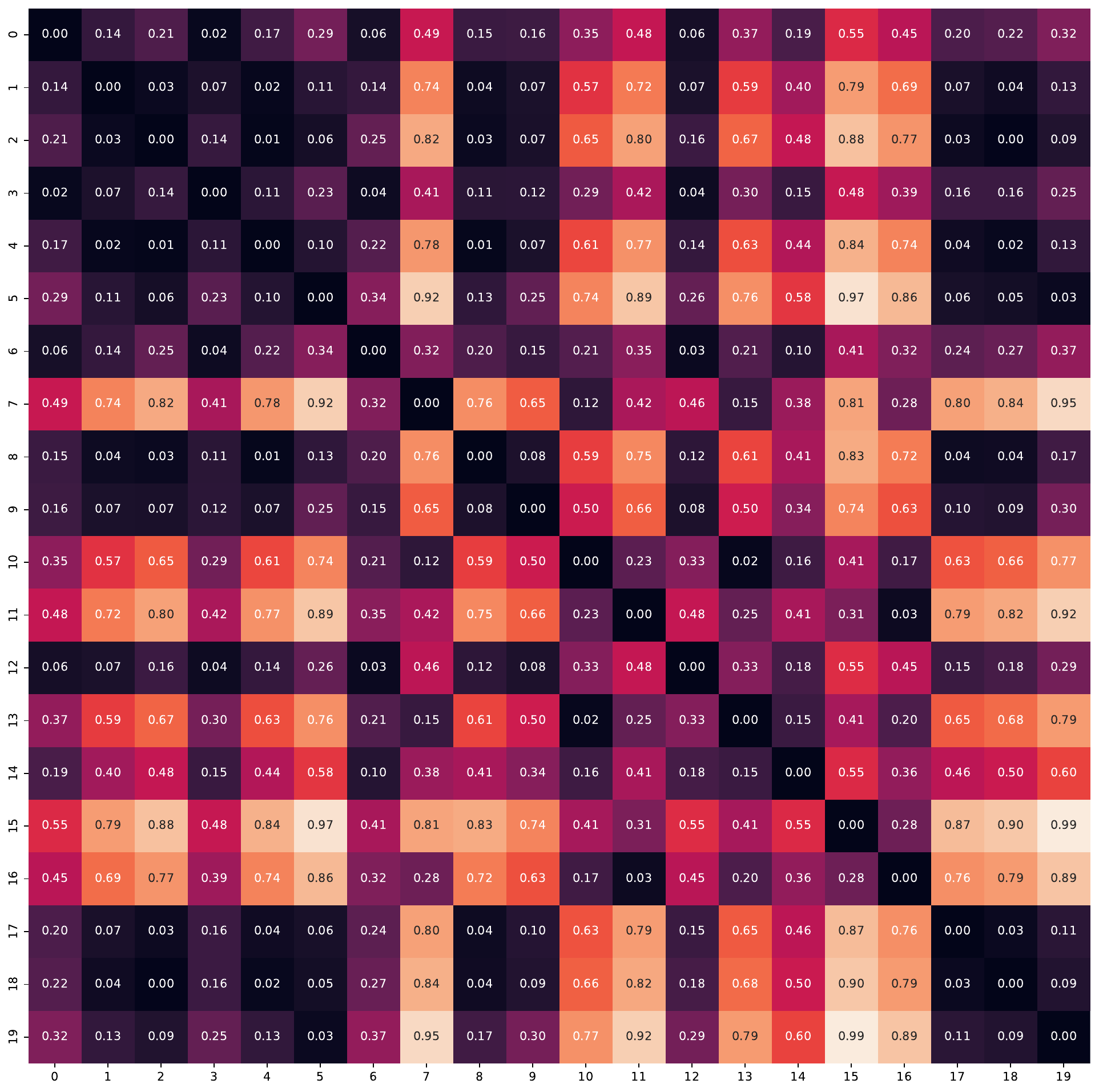}
		\caption{Top feature in COIL-20.}
	\end{subfigure}
	\begin{subfigure}[t]{0.32\textwidth}
		\includegraphics[width=\textwidth]{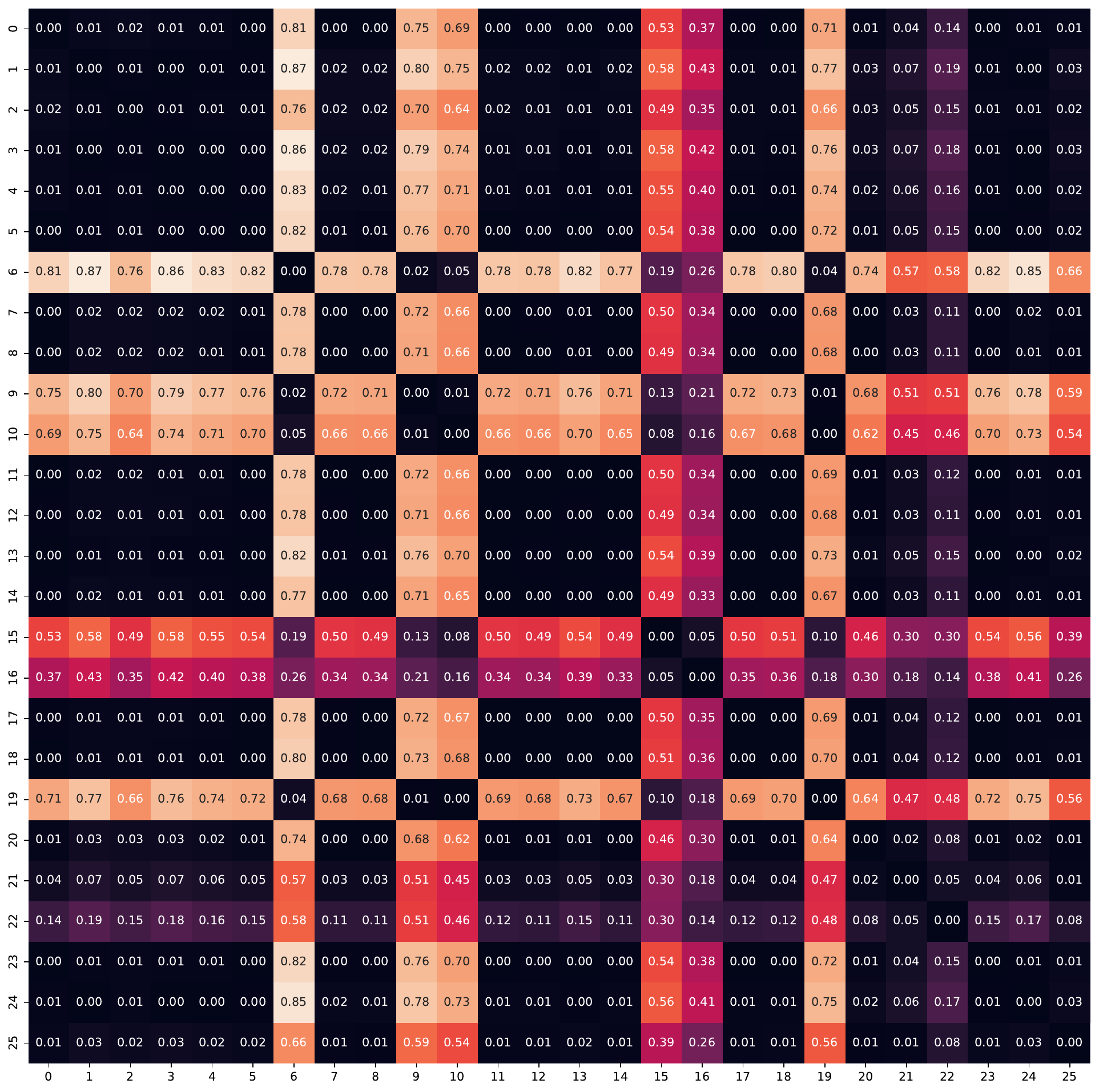}
		\caption{Top feature in ISOLET.}
	\end{subfigure}
	\begin{subfigure}[t]{0.32\textwidth}
		\includegraphics[width=\textwidth]{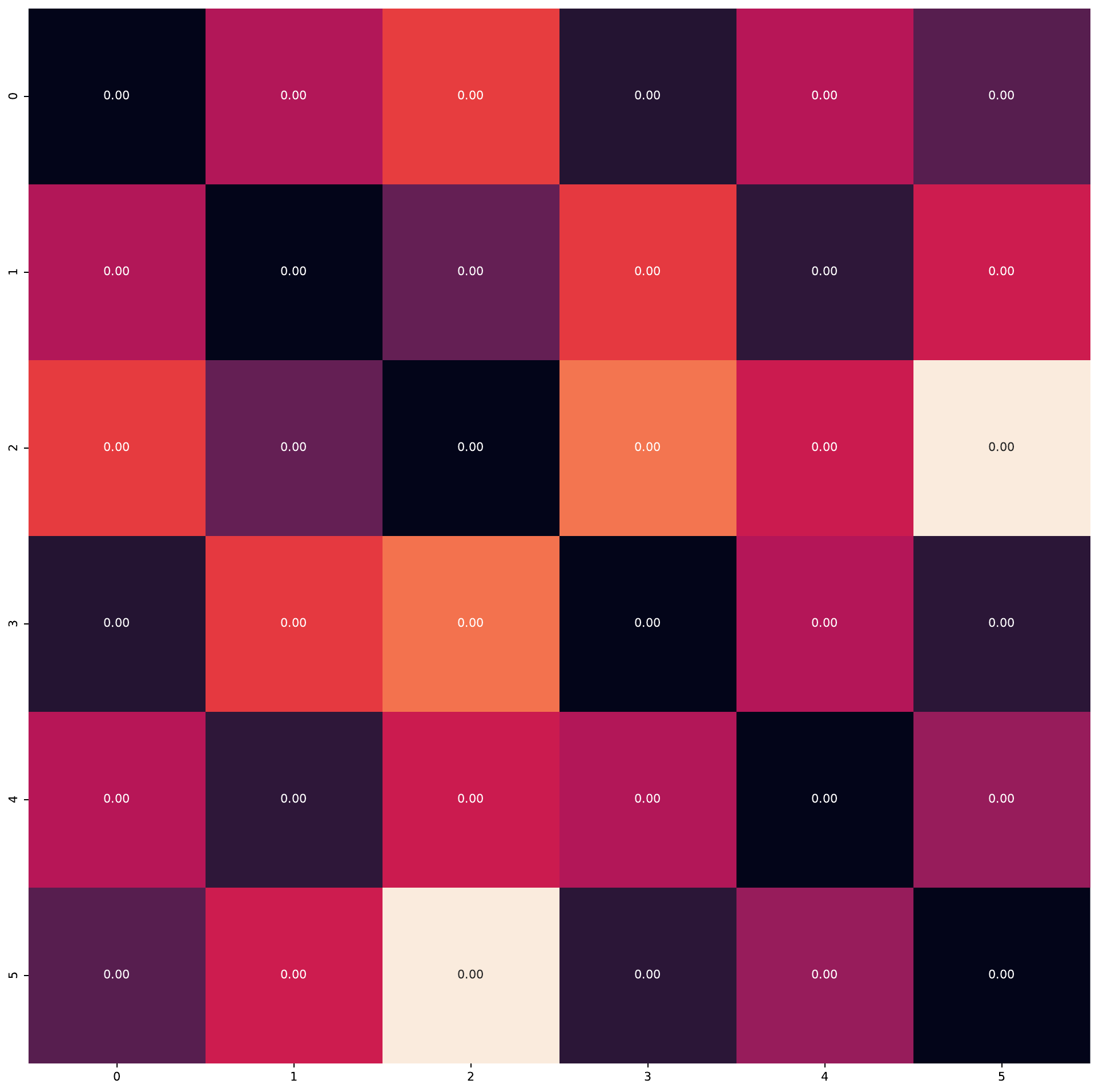}
		\caption{Least feature in Activity.}
	\end{subfigure}
	\begin{subfigure}[t]{0.32\textwidth}
		\includegraphics[width=\textwidth]{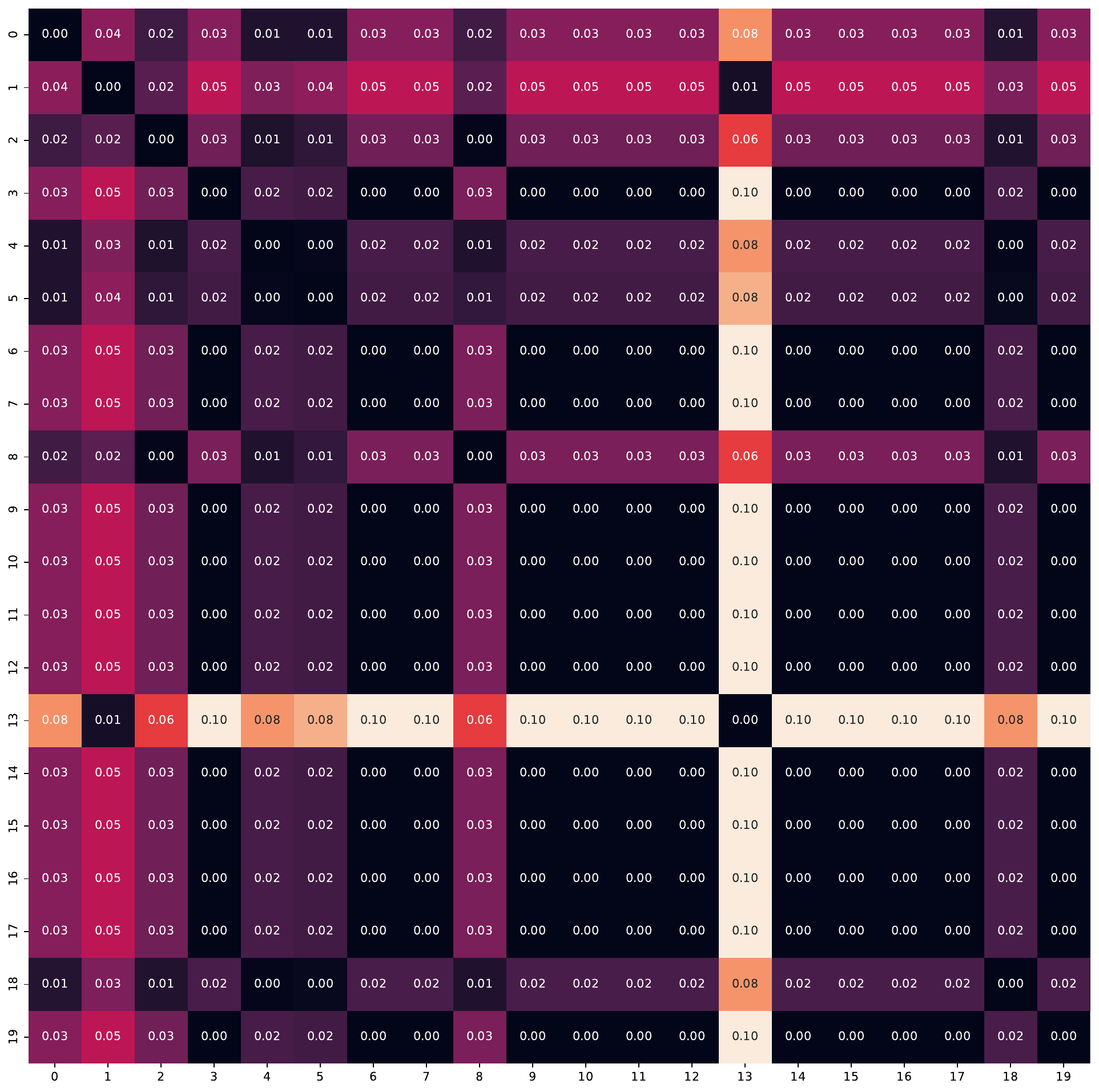}
		\caption{Least feature in COIL-20.}
	\end{subfigure}
	\begin{subfigure}[t]{0.32\textwidth}
		\includegraphics[width=\textwidth]{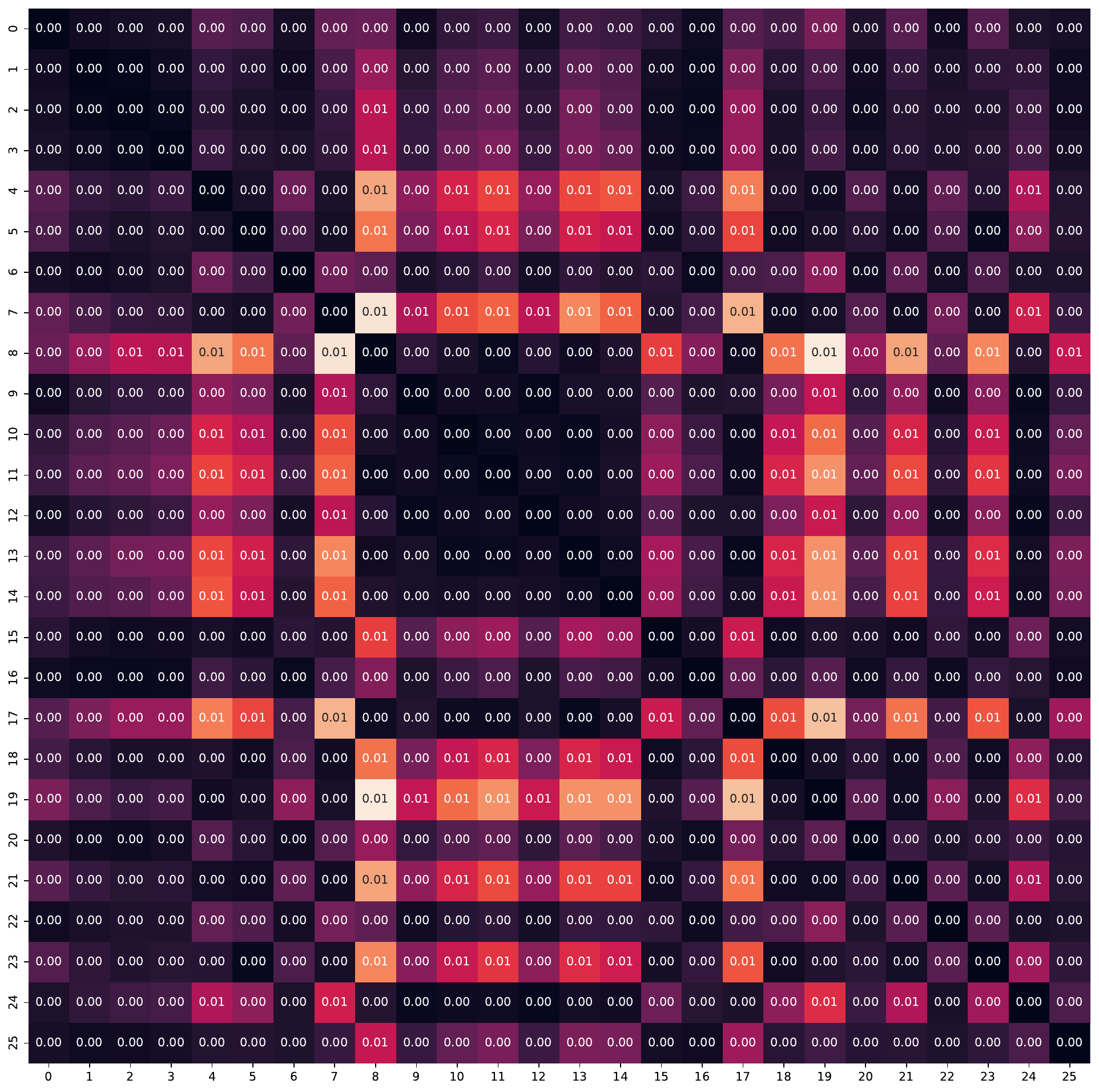}
		\caption{Least feature in ISOLET.}
	\end{subfigure}
	
	\begin{subfigure}[t]{0.32\textwidth}
		\includegraphics[width=\textwidth]{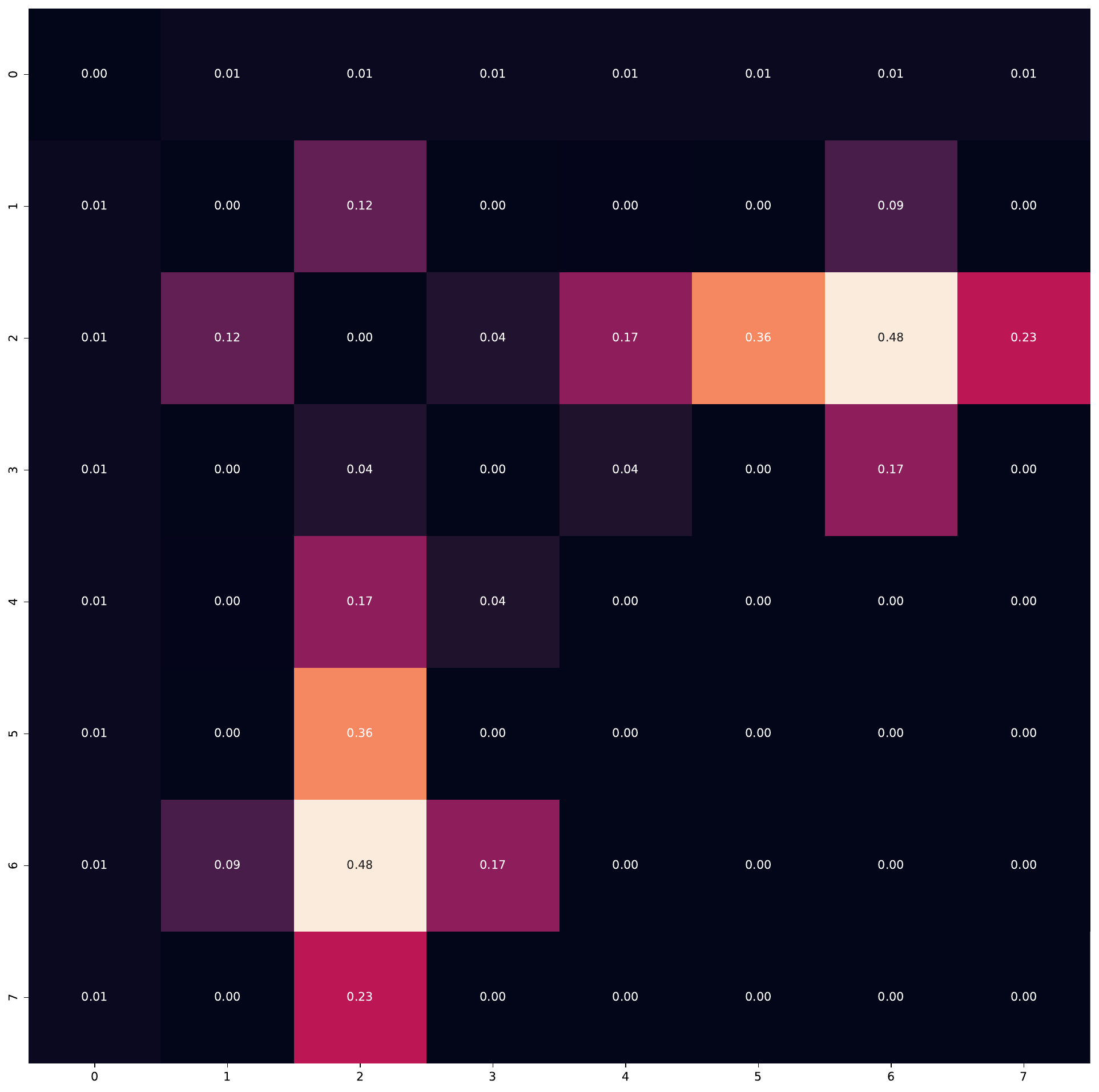}
		\caption{Top feature in MICE.}
	\end{subfigure}
	\begin{subfigure}[t]{0.32\textwidth}
		\includegraphics[width=\textwidth]{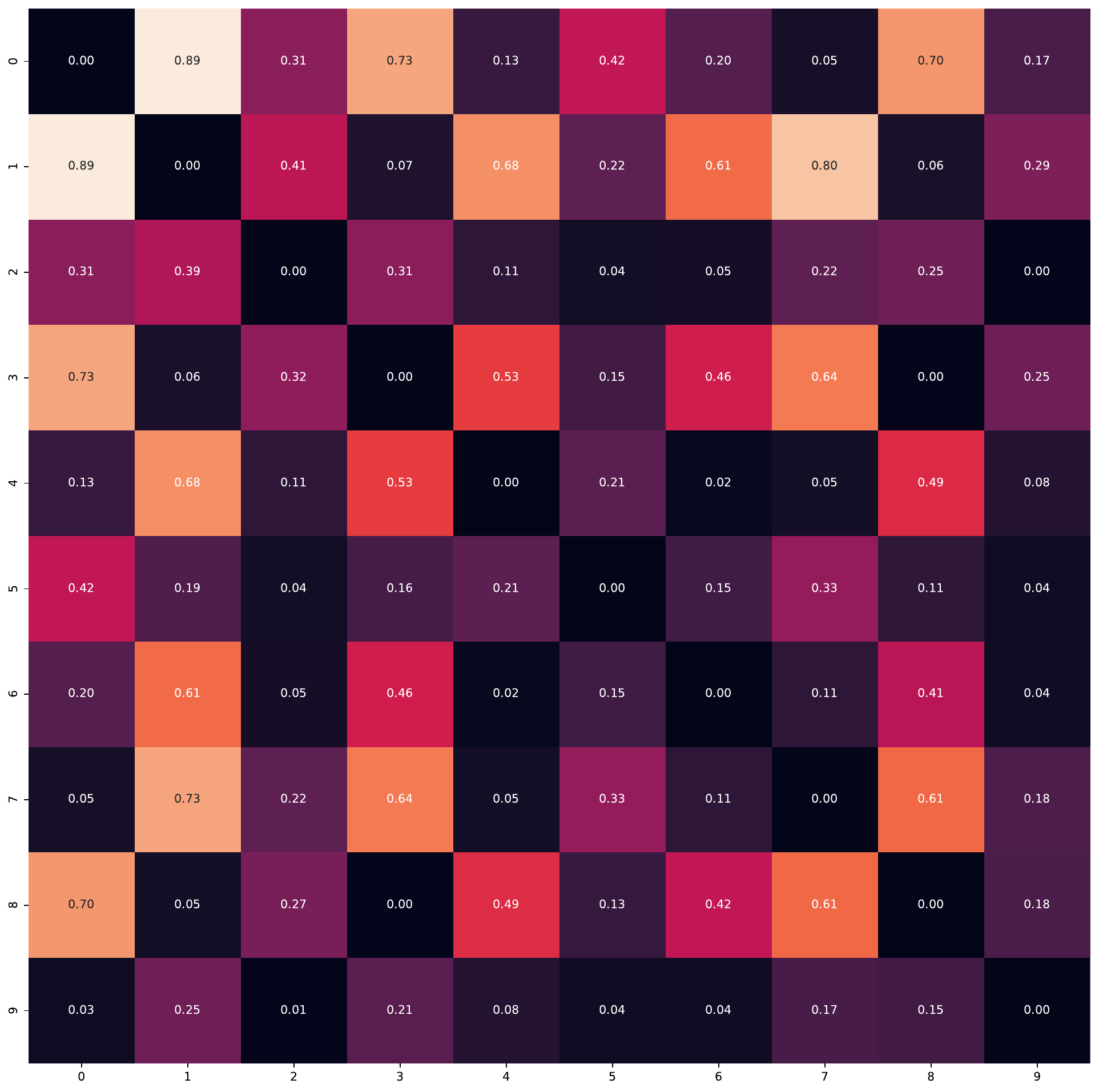}
		\caption{Top feature in MNIST.}
	\end{subfigure}
	\begin{subfigure}[t]{0.32\textwidth}
		\includegraphics[width=\textwidth]{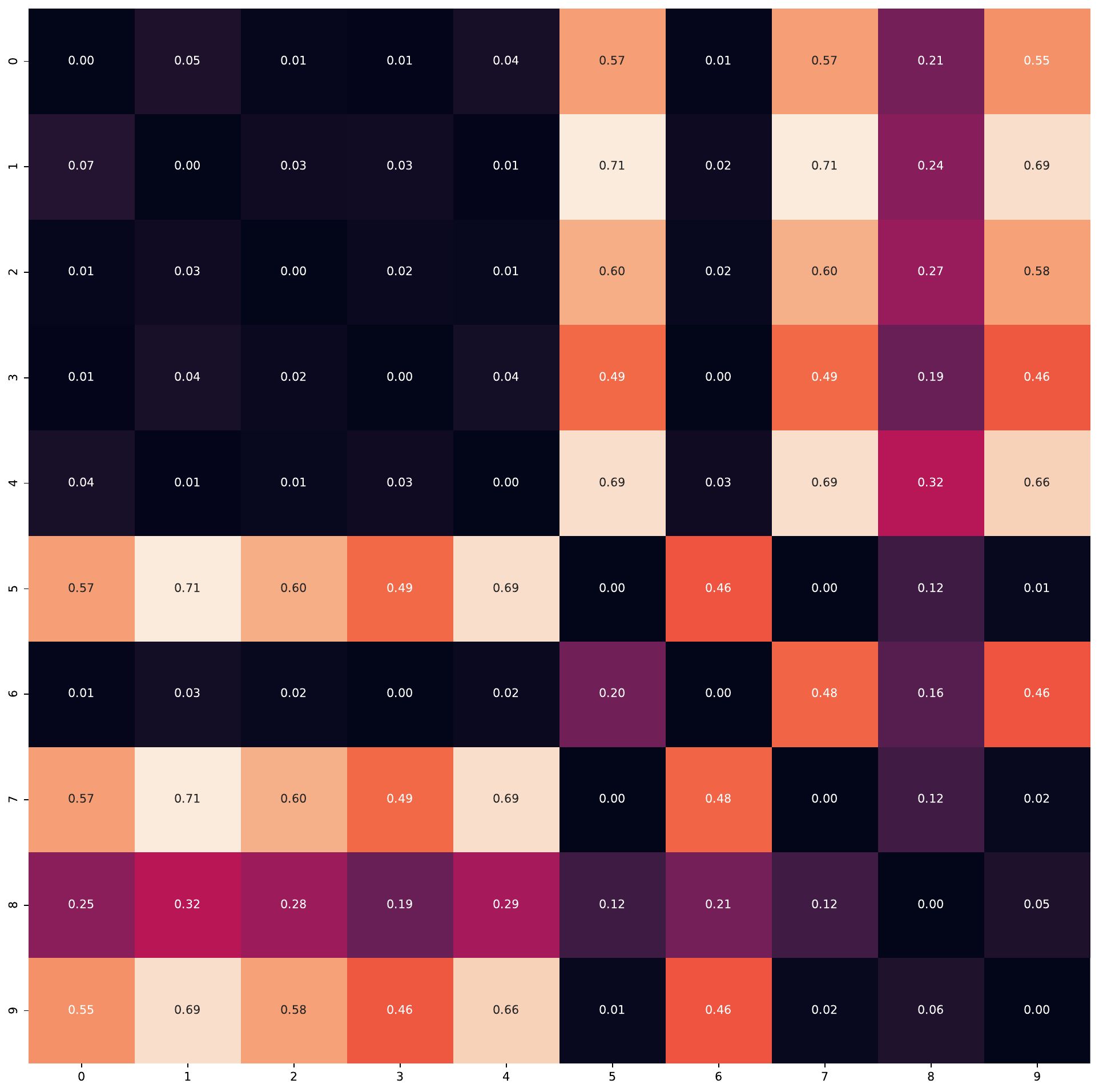}
		\caption{Top feature in F-MNIST.}
	\end{subfigure}
	\begin{subfigure}[t]{0.32\textwidth}
		\includegraphics[width=\textwidth]{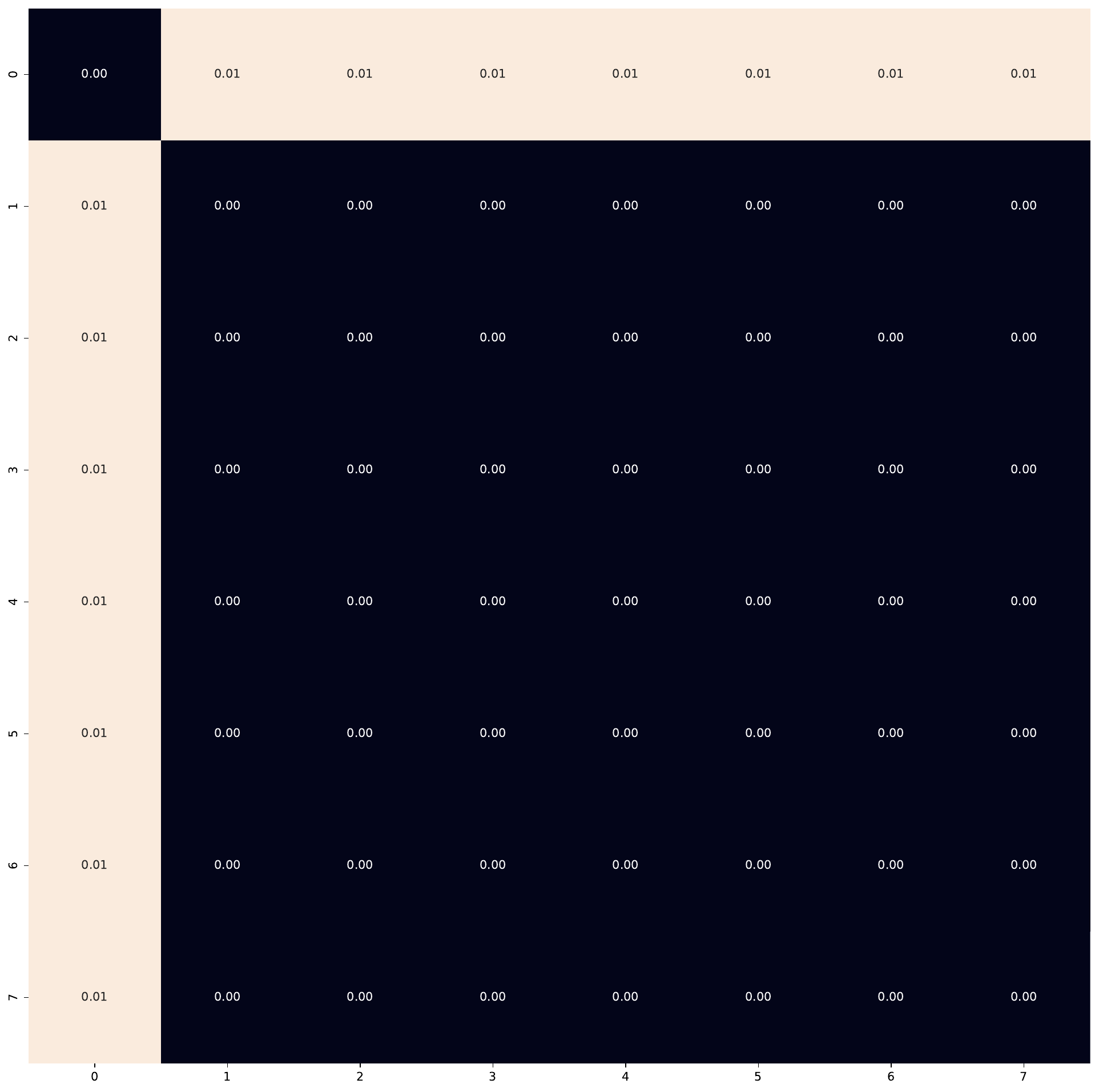}
		\caption{Least feature in MICE.}
	\end{subfigure}
	\begin{subfigure}[t]{0.32\textwidth}
		\includegraphics[width=\textwidth]{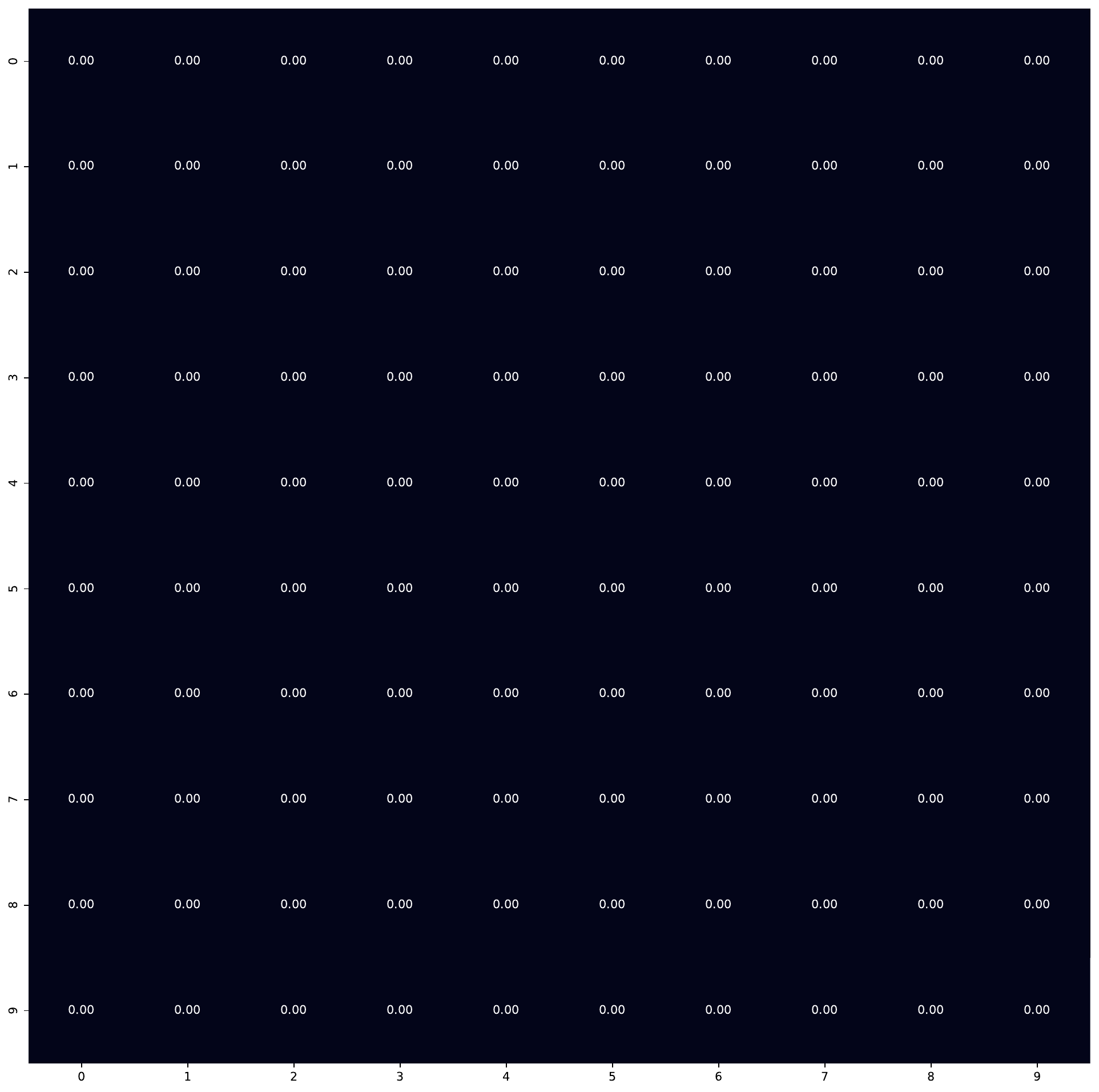}
		\caption{Least feature in MNIST.}
	\end{subfigure}
	\begin{subfigure}[t]{0.32\textwidth}
		\includegraphics[width=\textwidth]{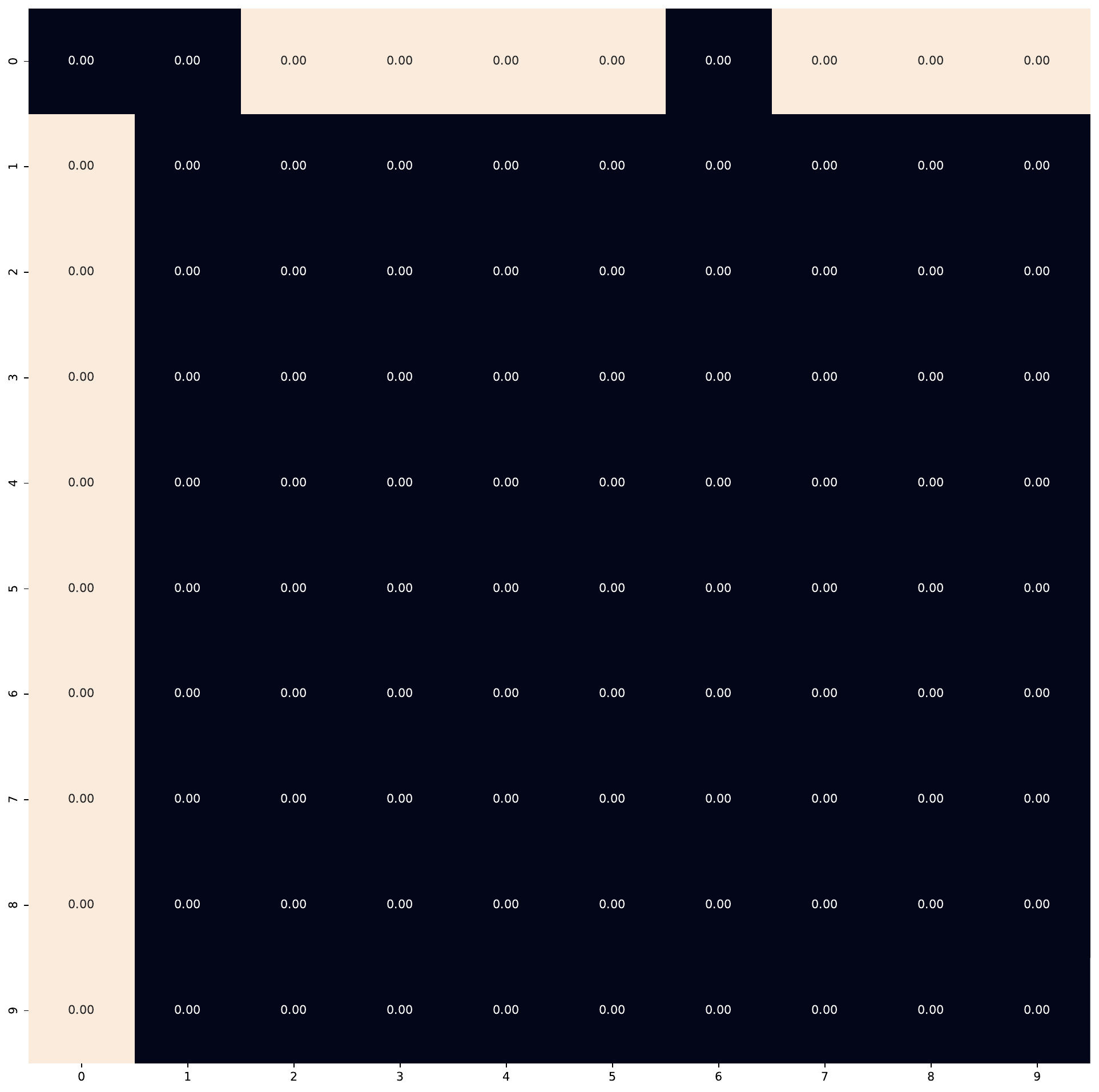}
		\caption{Least feature in F-MNIST.}
	\end{subfigure}
	
	\caption{Insights provided by the distance matrices.}
	\label{dist_mat_top_lst_single_feature}
\end{figure}

\subsection{Datasets}
We use benchmark datasets from various domains, such as image data, voice data, sensor data, and biomedical data. These datasets have been widely adopted for evaluating feature selection methods in the literature \cite{balin2019concrete, lemhadri2021lassonet}. We perform experiments on these datasets to evaluate the performance and robustness of our methods in comparison with existing methods. Table \ref{details_of_datasets} provides an overview of the datasets used in our experiments.

The datasets include:
\begin{itemize}
\item COIL-20: This dataset contains 1,440 grayscale images of 20 objects, each taken from 72 different angles.
\item MNIST: This dataset comprises 60,000 handwritten digits for training and 10,000 for testing. It is a common benchmark for image processing and machine learning.
\item Fashion-MNIST(F-MNIST): This dataset consists of 70,000 images of fashion items and is designed as a more difficult and realistic version of the original MNIST dataset.
\item nMNIST-AWGN: This is a variant of MNIST, generated by adding white Gaussian noise to the original images to mimic noisy environments. \item Human Activity Recognition (Activity): This dataset contains numerical items collected by sensors in a smartphone. Each item represents one of six activities of the user.
\item ISOLET: This dataset involves classifying spoken letters of the English alphabet. The number of samples are features derived from voice data.
\item MICE: This dataset comprises gene expression data about 77 proteins from 1080 mice with different genetic backgrounds and treatments.
\end{itemize}

We follow the same pre-processing step as in \cite{lemhadri2021lassonet} for the COIL-20 dataset, by resizing the images to 20*20 pixels, which reduces the number of features to 400.

\begin{table}[!ht]
	\centering
	\begin{adjustbox}{max width=\textwidth,keepaspectratio} 
		\begin{tabular}{|c|c|c|c|c|c|c|c|}
			\hline
			Dataset & Activity & COIL-20 & ISOLET & MICE & MNIST & Fashion-MNIST & nMNIST-AWGN \\ \hline
			\#(Features) & 561 & 400 & 617 & 77 & 784 & 784 & 784 \\ \hline
			\#(Samples) & 5744 & 1440 & 7797 & 1080 & 70000 & 70000 & 70000 \\ \hline
			\#(Classes) & 6 & 20 & 26 & 8 & 10 & 10 & 10 \\ \hline
		\end{tabular}
	\end{adjustbox}
	\caption{The Details of Benchmark Datasets}
	\label{details_of_datasets}
\end{table}
We randomly divide each data set into training and test sets. To ensure a fair comparison, the training and test sets are the same for all methods, using the same random seed. For the COIL-20, MICE, and ISOLET datasets, which have relatively few samples per class, we use a 70-30 split for training and testing. For the MNIST, Fashion-MNIST, and nMNIST-AWGN datasets, which have many samples, we randomly sample a subset of them for training and testing with a 20-80 split to avoid memory overflow on our PC.

\subsection{Methodology}
To evaluate the effectiveness of our methods, we conduct experiments from the following perspectives: 
\begin{itemize} 
	\item We show that our methods are able to identify features that are consistent across different learning paradigms, such as supervised and unsupervised settings. 
	\item We show that our methods can effectively filter out features that are irrelevant or redundant to the learning task, regardless of the availability of labels. 
	\item We analyze how the degree of variation among features affects classification performance. 
	\item We compare the similarity between the original and transformed feature sets and examine its impact on accuracy. 
\end{itemize}

We use two types of datasets to test our methods: benchmark datasets and synthetic noise data. The synthetic noise data have similar low-order statistical properties to real-world datasets. We apply our methods to both types of datasets for the first two evaluation criteria, and only to the benchmark datasets for the last two criteria. We compare our methods with existing feature selection methods, both filter and embedded, that are based on correlation measures. The filter methods include JMI \cite{yang1999data}, CFS \cite{hall1999feature}, Fisher Score \cite{duda2000pattern}, Trace-ratio \cite{nie2008trace}, MRMR \cite{hanchuanpeng2005feature}, CMIM \cite{fleuret2004fast}, NDFS \cite{li2012unsupervised}, and UDFS \cite{yang2011l2}. The embedded methods include HSIC-Lasso \cite{yamada2014highdimensional} and LassoNet \cite{lemhadri2021lassonet}. We use the scikit-feature package \cite{li2018feature} to implement the filter methods, the pyHSIC package \cite{yamada2014highdimensional} \cite{climente-gonzalez2019block} to implement HSIC-Lasso, and the official source code to implement LassoNet. We use the default settings for the hyperparameters of all the compared methods.

For fairness, we feed the feature sets selected by different methods as input to an independent classifier to explore the effectiveness of each method. We run each feature selection method in comparison to select a varying number of features, and measure the accuracy obtained by downstream classifiers when fed with these features as the metric to quantify the performance of feature selection methods. We report the average results of these methods from 10 runs on different training sets. To ensure fairness, we randomly split the dataset and ensure that all methods work on the same data in each run.

\begin{figure}[!ht]
	\centering
	\begin{subfigure}[t]{0.32\textwidth}
		\includegraphics[width=\textwidth]{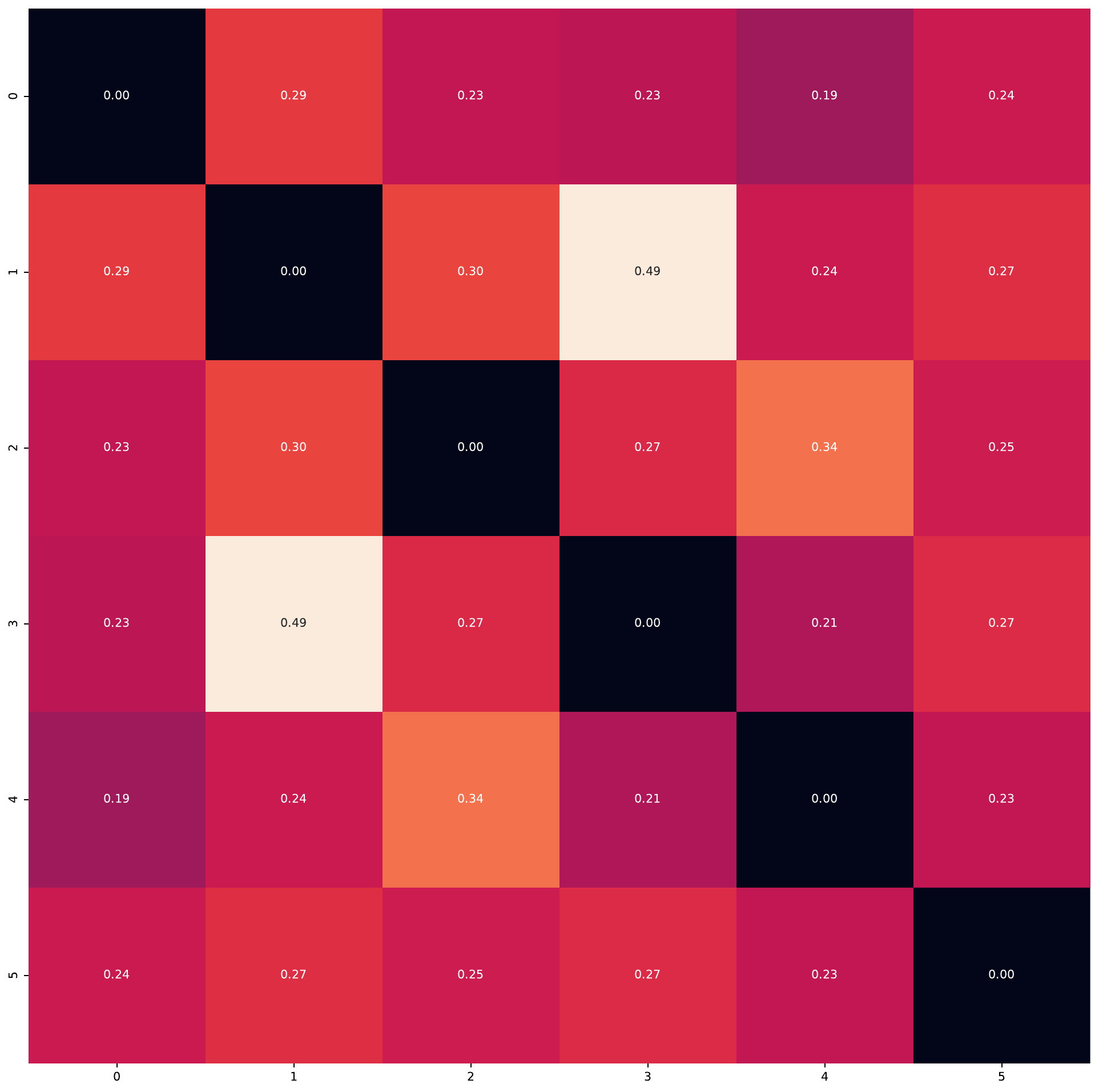}
		\caption{Top feature in Activity.}
	\end{subfigure}
	\begin{subfigure}[t]{0.32\textwidth}
		\includegraphics[width=\textwidth]{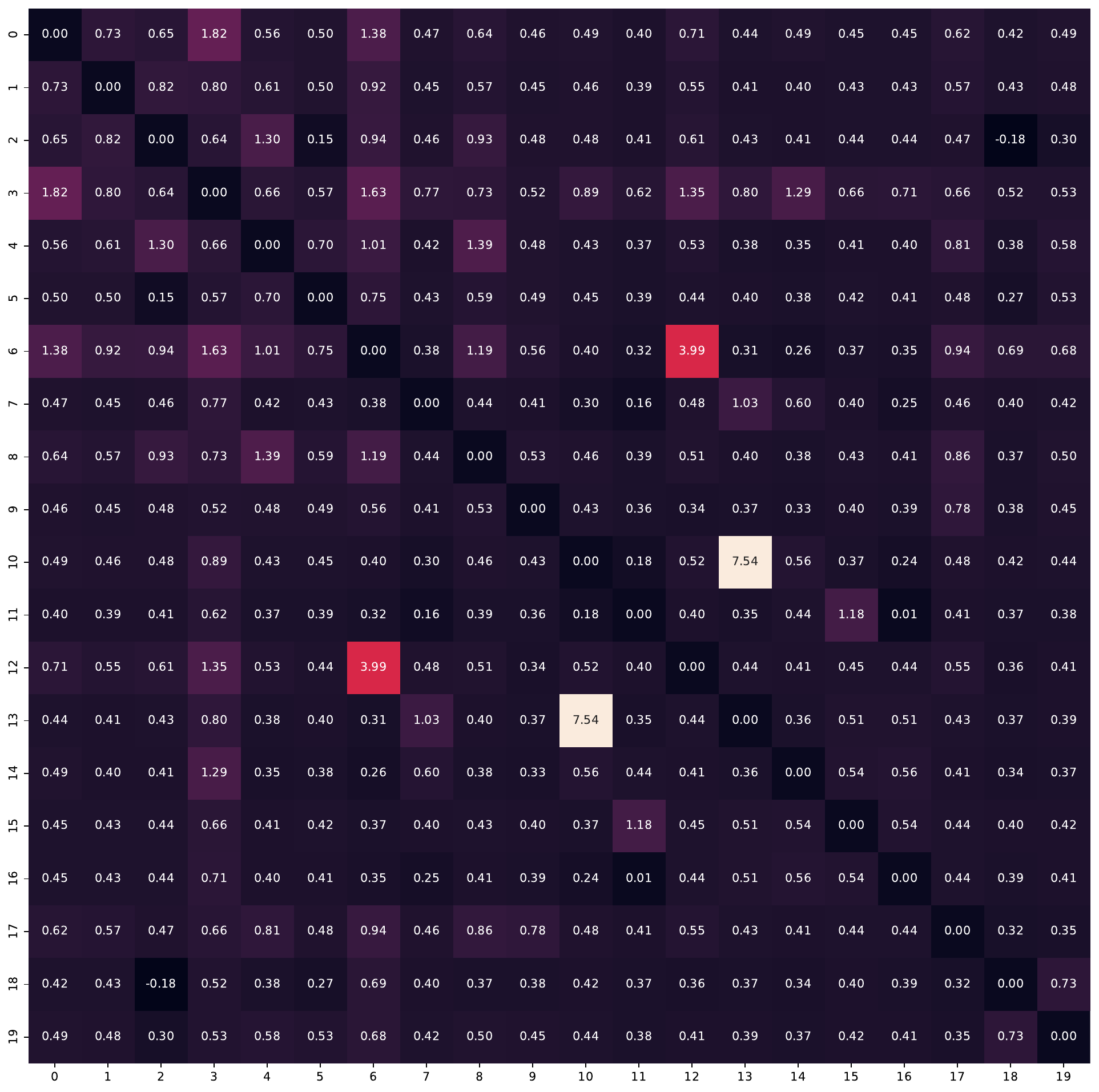}
		\caption{Top feature in COIL-20.}
	\end{subfigure}
	\begin{subfigure}[t]{0.32\textwidth}
		\includegraphics[width=\textwidth]{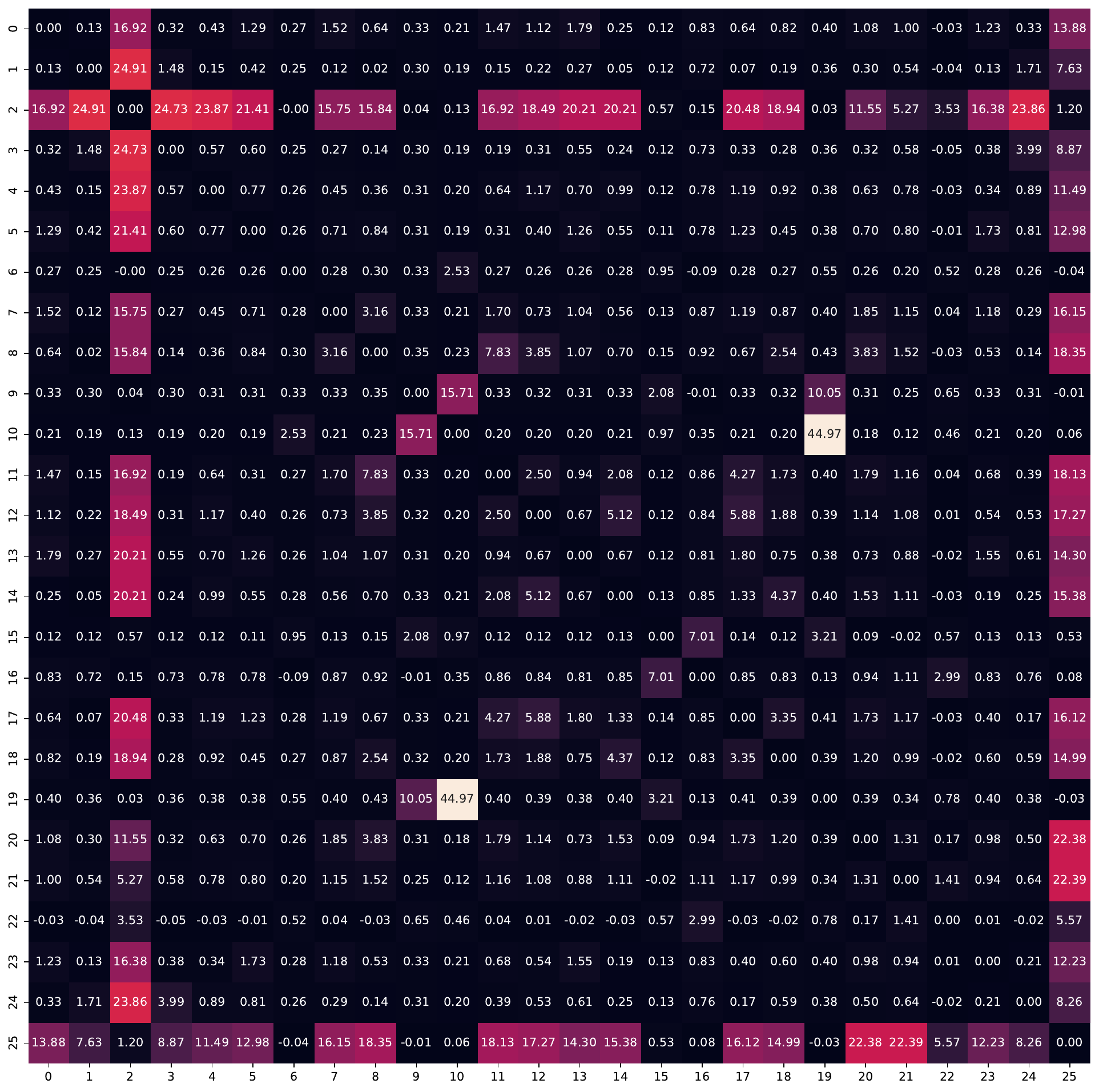}
		\caption{Top feature in ISOLET.}
	\end{subfigure}
	\begin{subfigure}[t]{0.32\textwidth}
		\includegraphics[width=\textwidth]{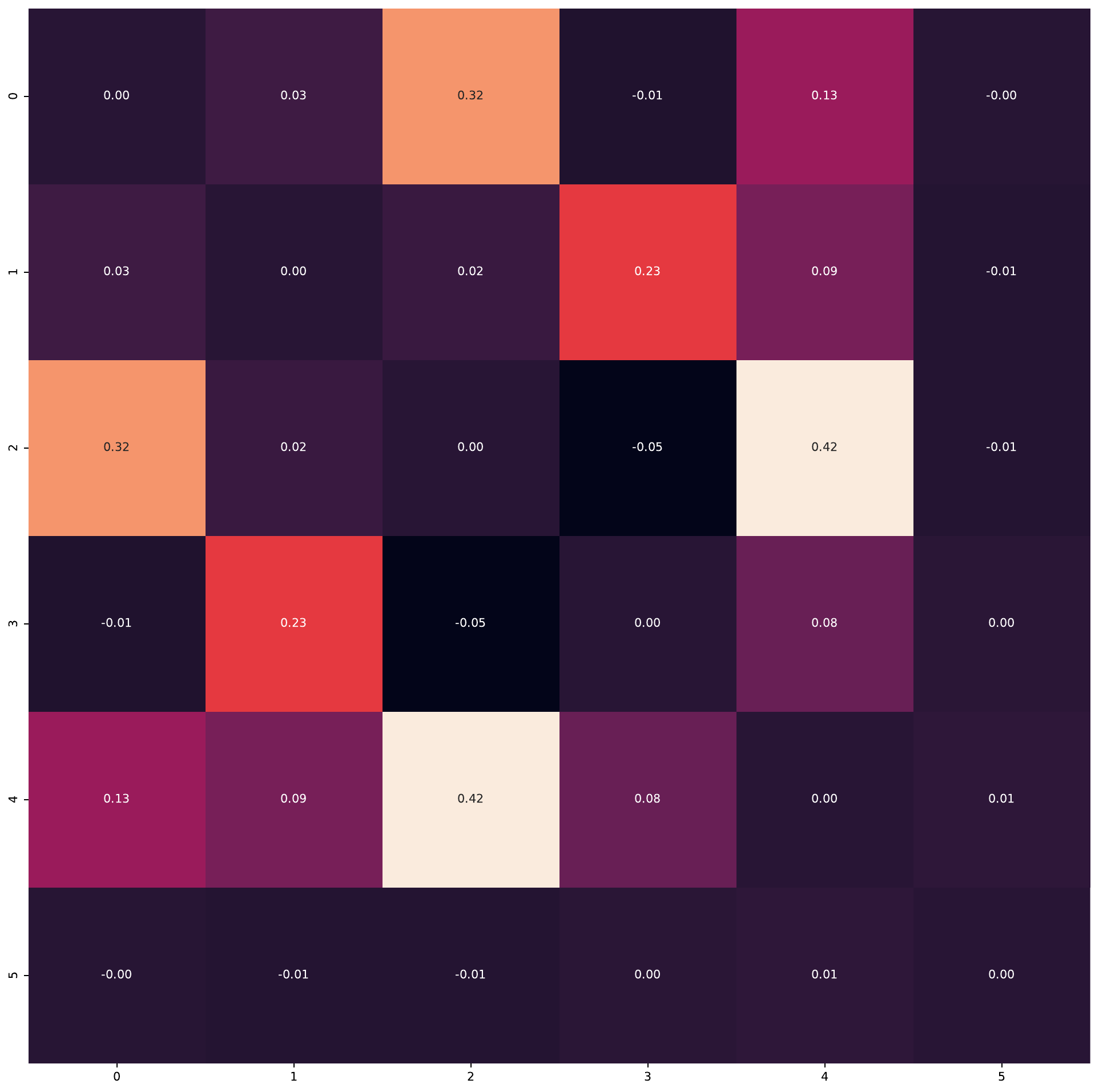}
		\caption{Least feature in Activity.}
	\end{subfigure}
	\begin{subfigure}[t]{0.32\textwidth}
		\includegraphics[width=\textwidth]{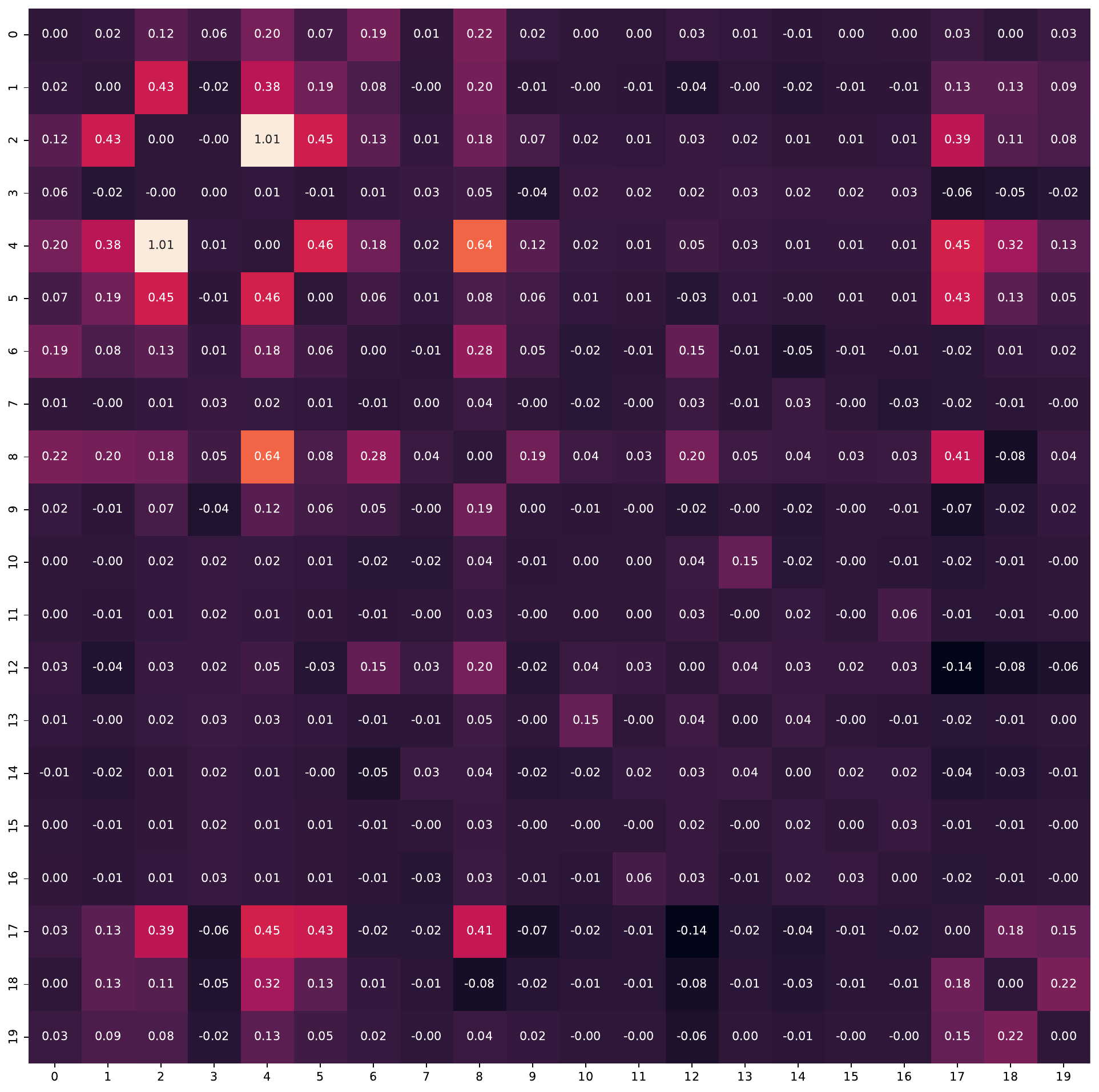}
		\caption{Least feature in COIL-20.}
	\end{subfigure}
	\begin{subfigure}[t]{0.32\textwidth}
		\includegraphics[width=\textwidth]{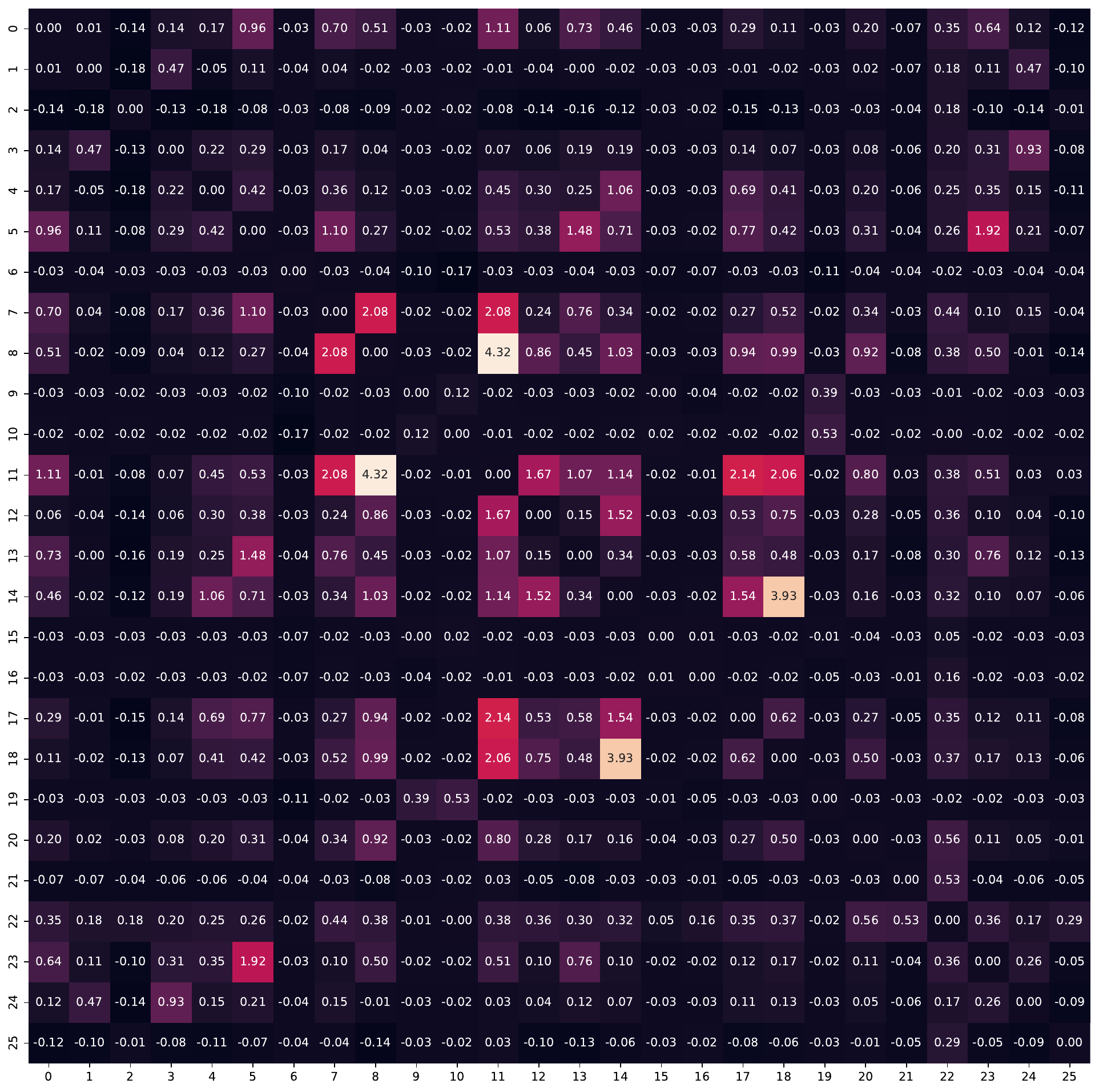}
		\caption{Least feature in ISOLET.}
	\end{subfigure}
	
	\begin{subfigure}[t]{0.32\textwidth}
		\includegraphics[width=\textwidth]{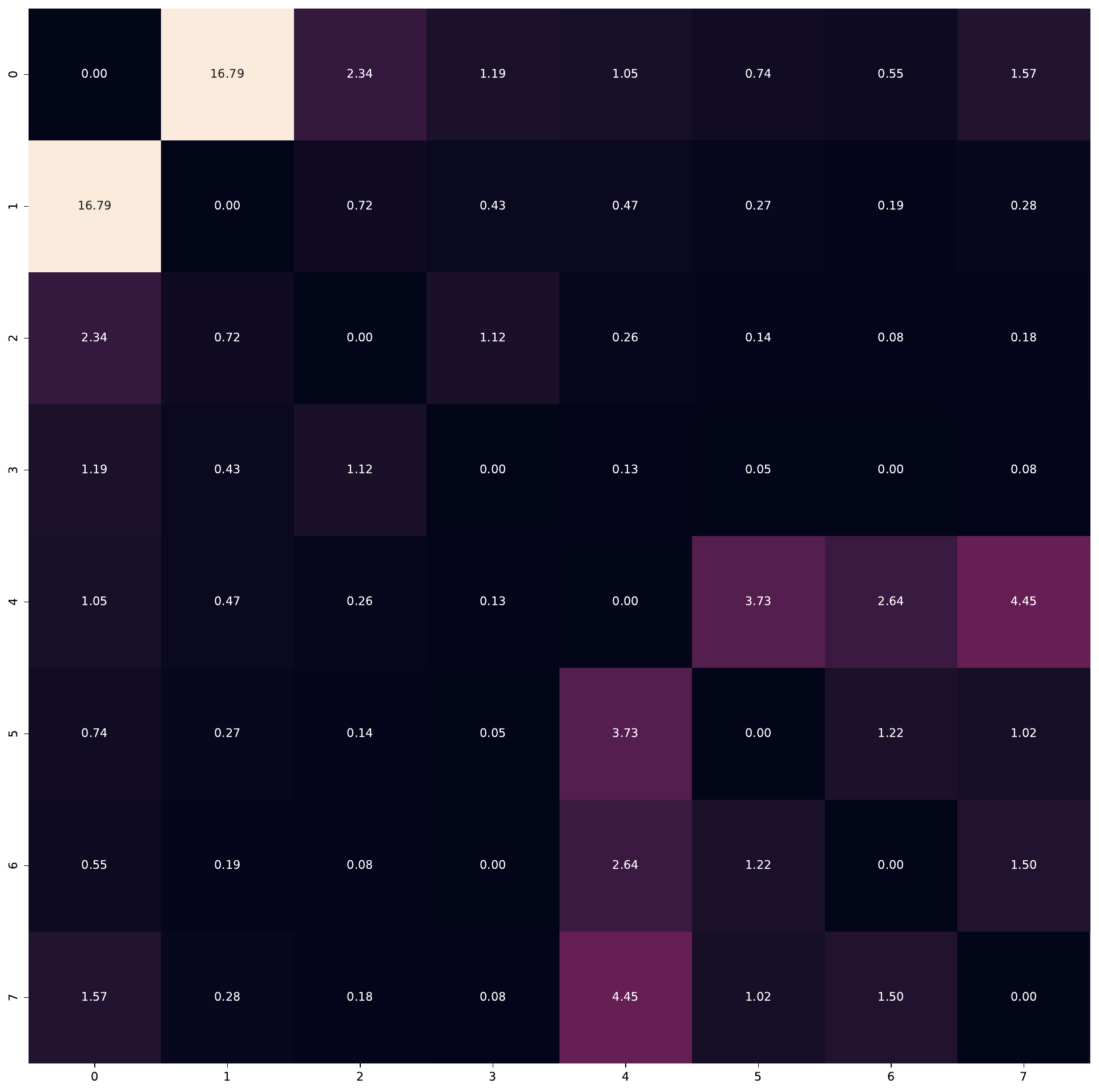}
		\caption{Top feature in MICE.}
	\end{subfigure}
	\begin{subfigure}[t]{0.32\textwidth}
		\includegraphics[width=\textwidth]{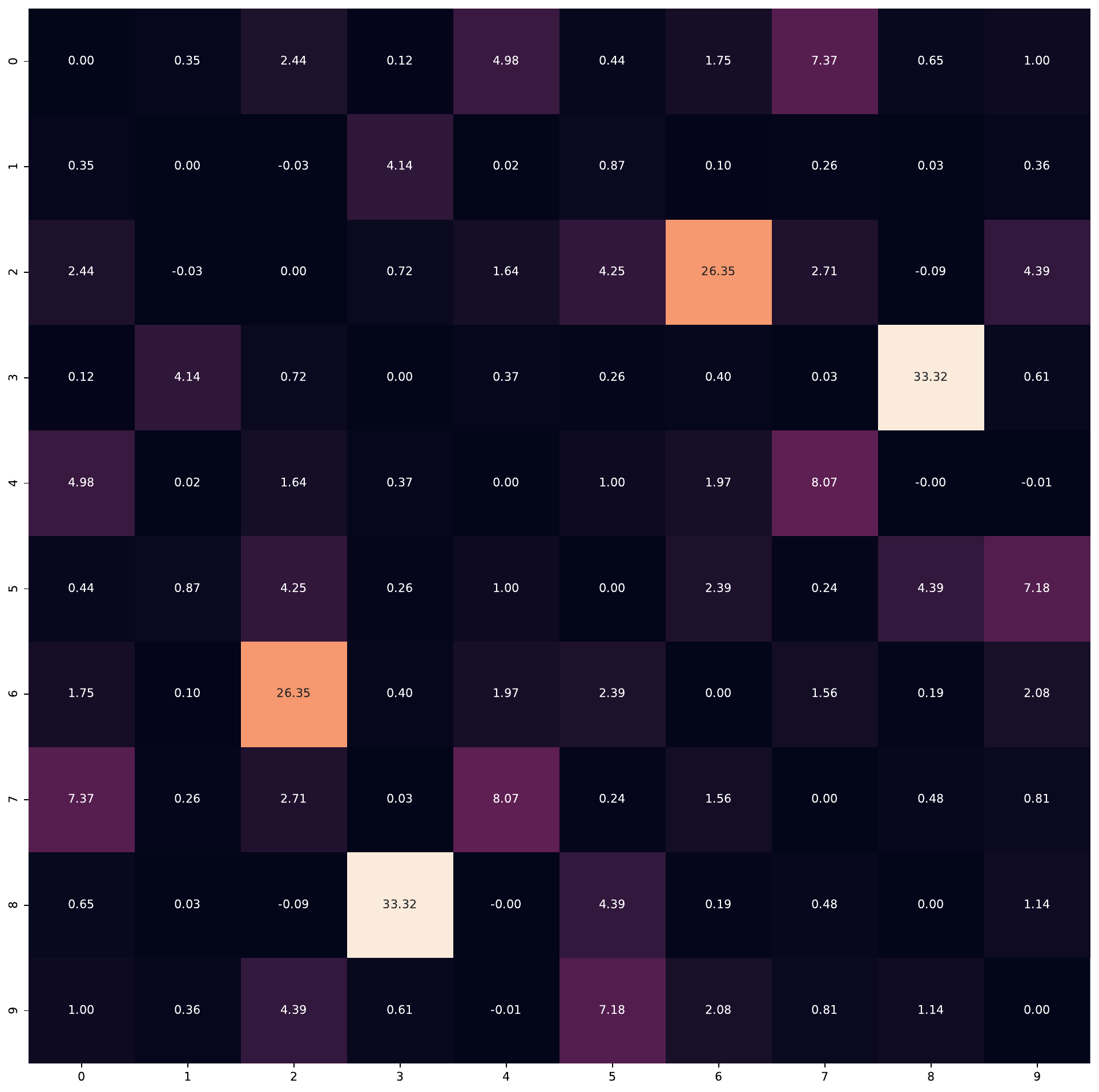}
		\caption{Top feature in MNIST.}
	\end{subfigure}
	\begin{subfigure}[t]{0.32\textwidth}
		\includegraphics[width=\textwidth]{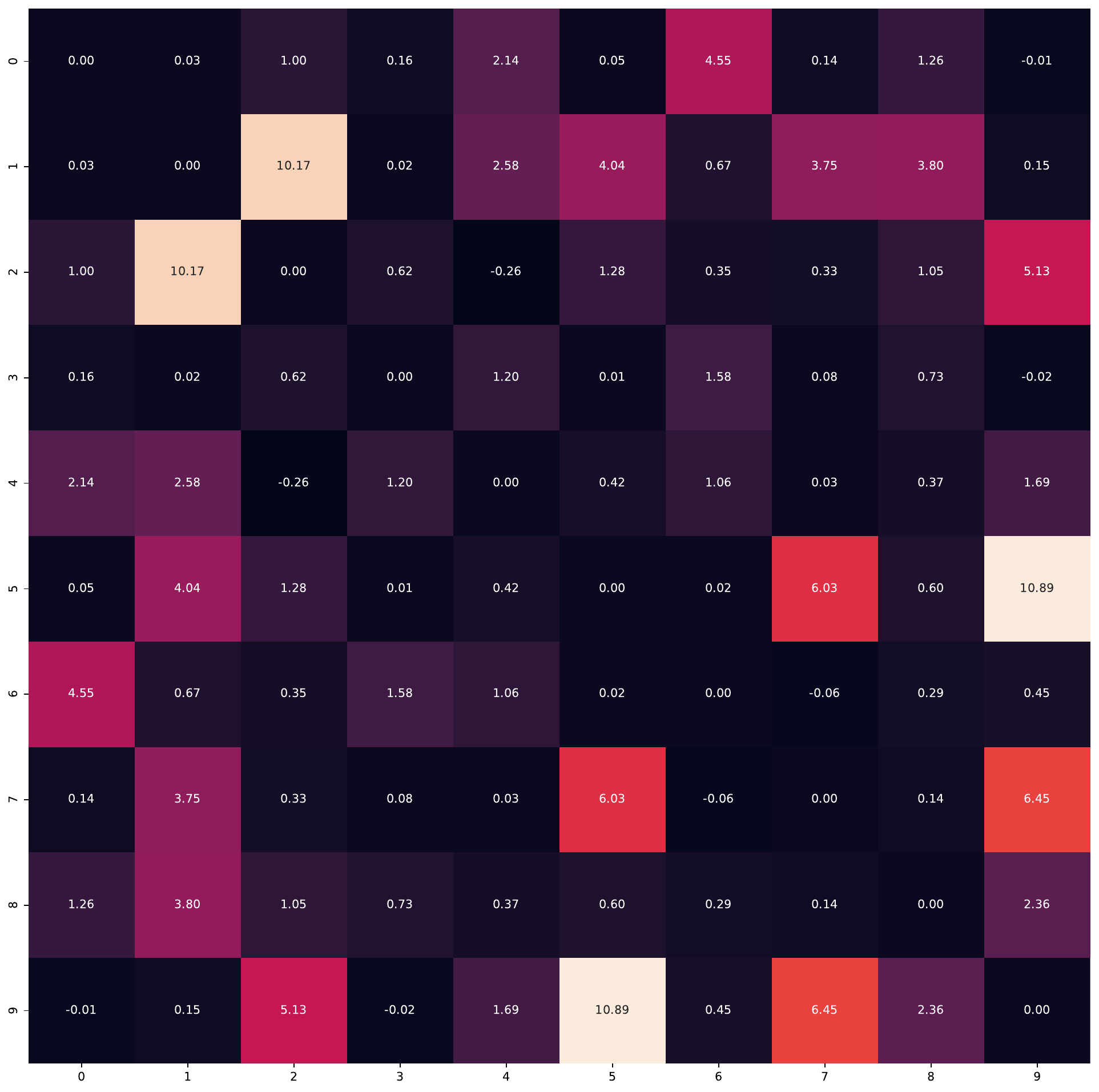}
		\caption{Top feature in F-MNIST.}
	\end{subfigure}
	\begin{subfigure}[t]{0.32\textwidth}
		\includegraphics[width=\textwidth]{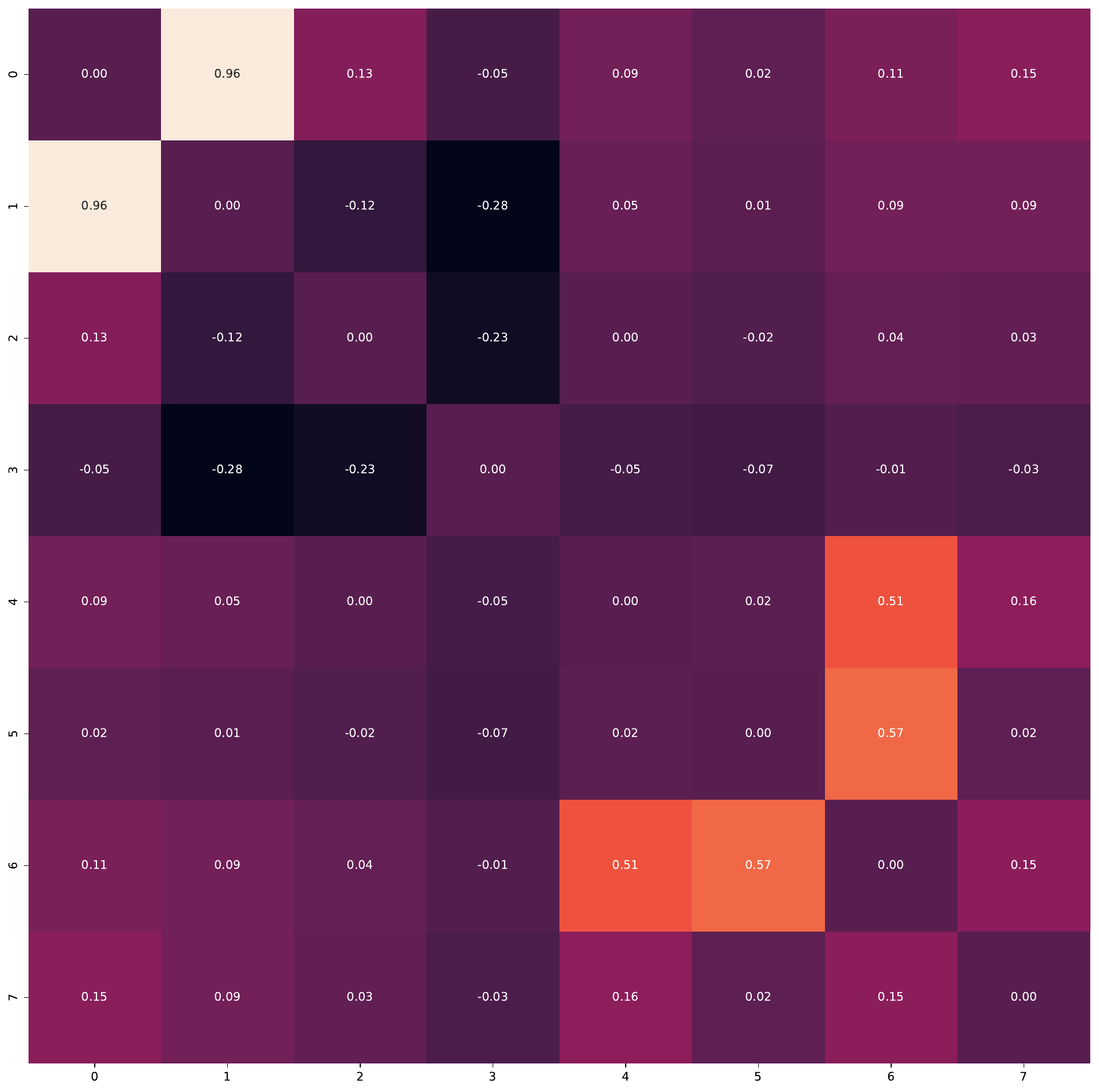}
		\caption{Least feature in MICE.}
	\end{subfigure}
	\begin{subfigure}[t]{0.32\textwidth}
		\includegraphics[width=\textwidth]{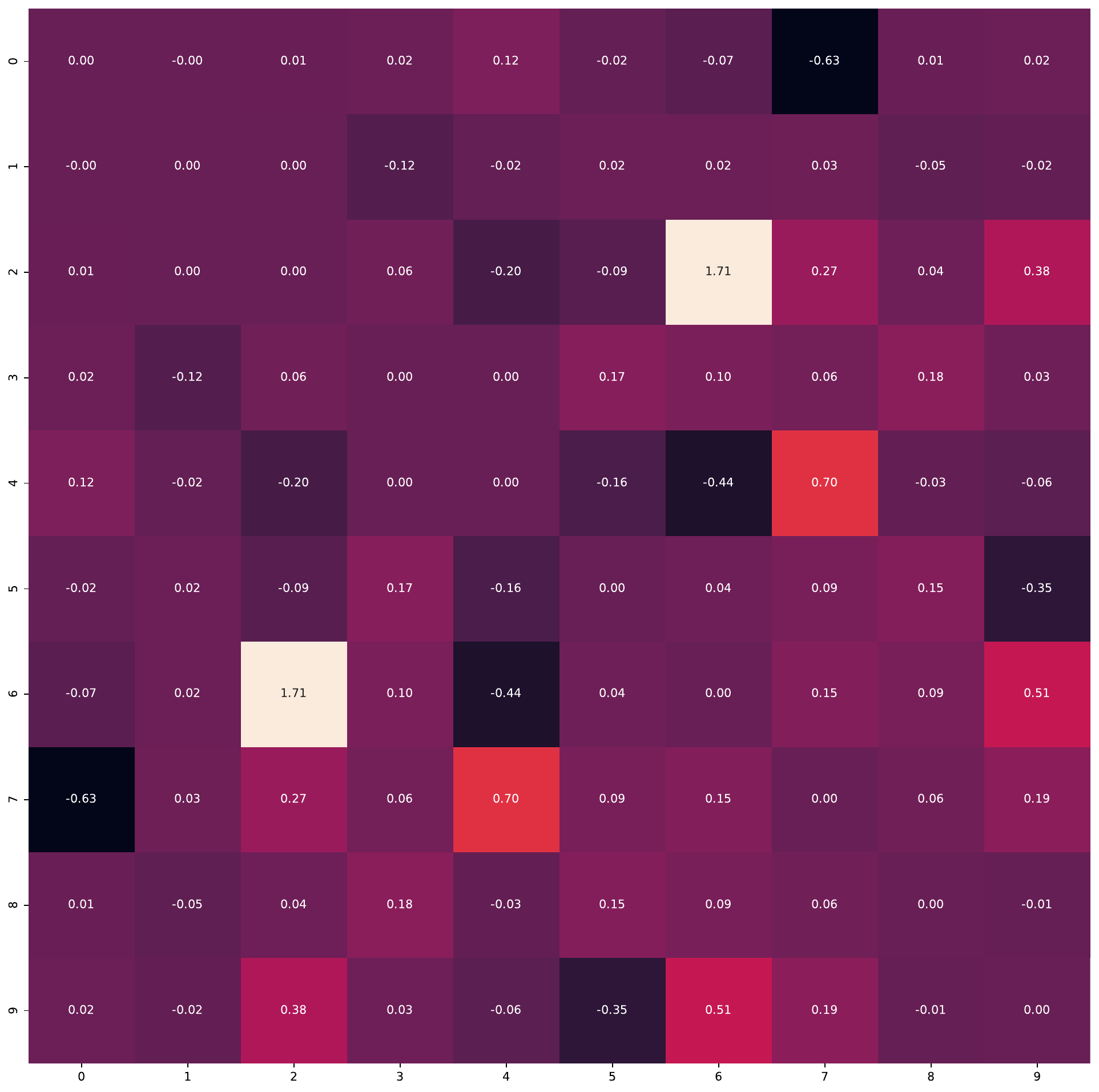}
		\caption{Least feature in MNIST.}
	\end{subfigure}
	\begin{subfigure}[t]{0.32\textwidth}
		\includegraphics[width=\textwidth]{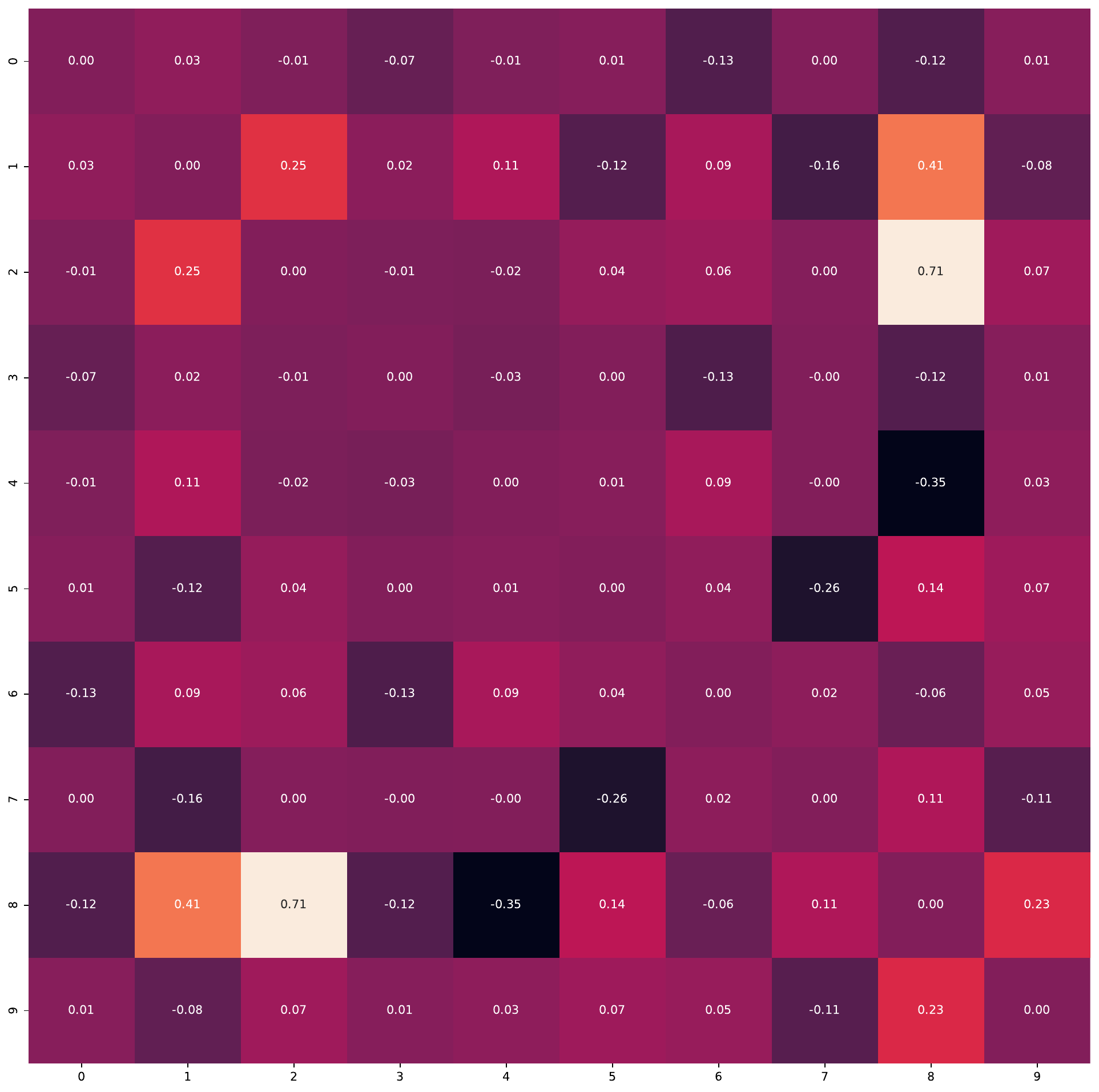}
		\caption{Least feature in F-MNIST.}
	\end{subfigure}
	
	\caption{Distance matrices of the relative change value brought by a single feature. This figure shows the relative change value due to the addition of the features with the largest (top) and smallest disparity, respectively.}
	\label{dist_mat_top_lst_influ_single_feature}
\end{figure}

\subsection{Experiment 1: when features are irrelevant}

In previous work, features in datasets have been categorized into relevant and irrelevant features \cite{yu2004efficient}. A common requirement for criteria in feature selection is to quantify the relevance of features.

First, we consider the supervised learning setting. We implement our methods on real-world benchmark datasets and present the distance matrix of features from these datasets. For each benchmark dataset, we select two features that have the largest and smallest disparity between classes, respectively. As shown in Fig \ref{dist_mat_top_lst_single_feature}, features with the largest disparity show significant superiority compared to their counterparts and features from synthetic datasets. Furthermore, the features with the largest disparity show more specific structures in the distance matrices than their counterparts, indicating that they may not be nearly as good at distinguishing all class pairs. This phenomenon also appears when compared to noise features, as shown in Fig \ref{dist_matrix_relev_noise_1fs}.

We also consider the influence of newly added features. As shown in Fig \ref{dist_mat_top_lst_influ_single_feature}, adding a least relevant feature to the selected feature set hardly increases the disparity across classes, while the top relevant feature significantly increases the disparity shown in the distance matrices.

In the unsupervised learning setting, we consider a feature set rather than a single feature. We consider the feature set containing top-relevant features, least-relevant features, and noise features, respectively, each feature set starts with size 5 and adds 5 features belongs to same type each step.
The Gromov-Wasserstein distances between the submatrices and the original datasets show a significant difference between these feature sets.
As can be seen in Fig \ref{increse_var_feature_vs_top50}, the feature set containing top-relevant features becomes more similar to original datasets under GWD, while the GWD between original data matrices and submatrices corresponding to sets consisting of least relevant features and noise features change little, that relevant features make the submatrices more closer to original data matrices.

\begin{figure}[!ht]
	\centering
	\begin{subfigure}[t]{0.99\textwidth}
		\includegraphics[width=\textwidth]{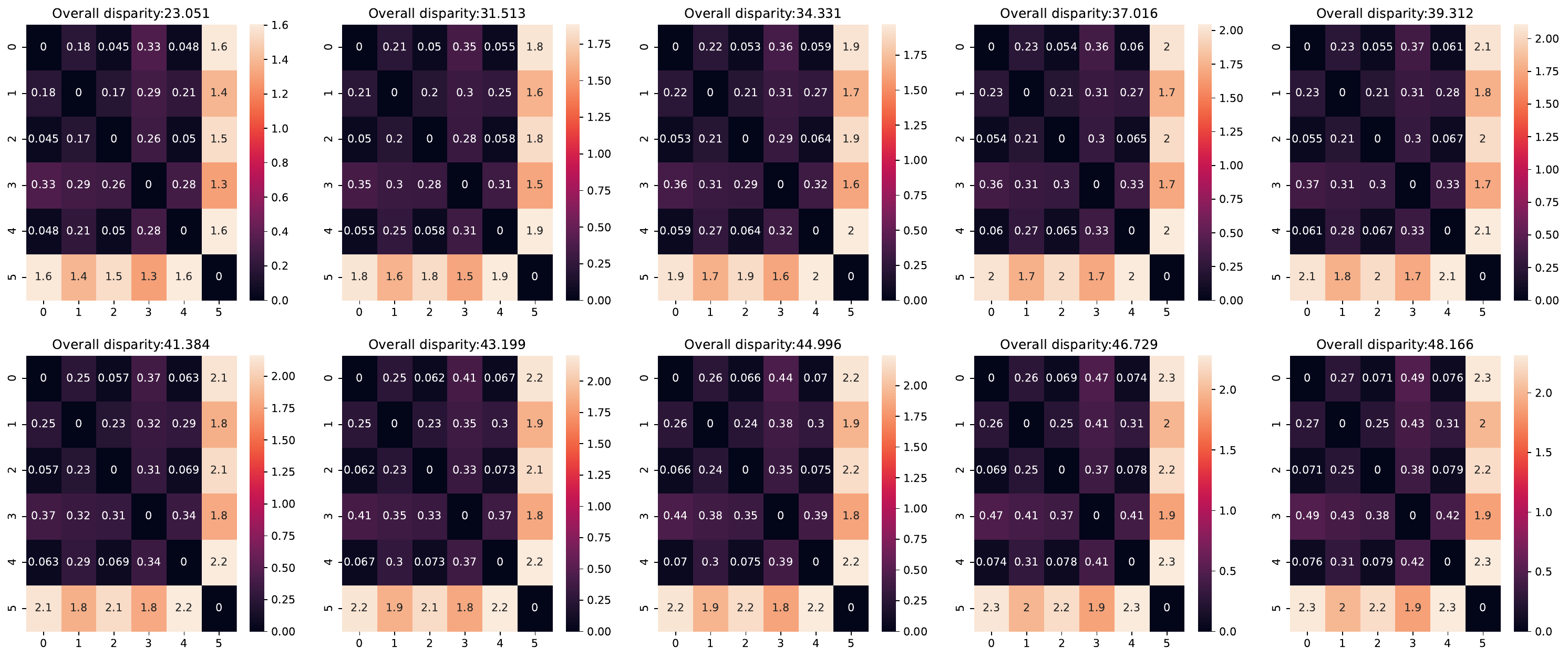}
		\caption{Top feature in Activity.}
	\end{subfigure}
	\begin{subfigure}[t]{0.99\textwidth}
		\includegraphics[width=\textwidth]{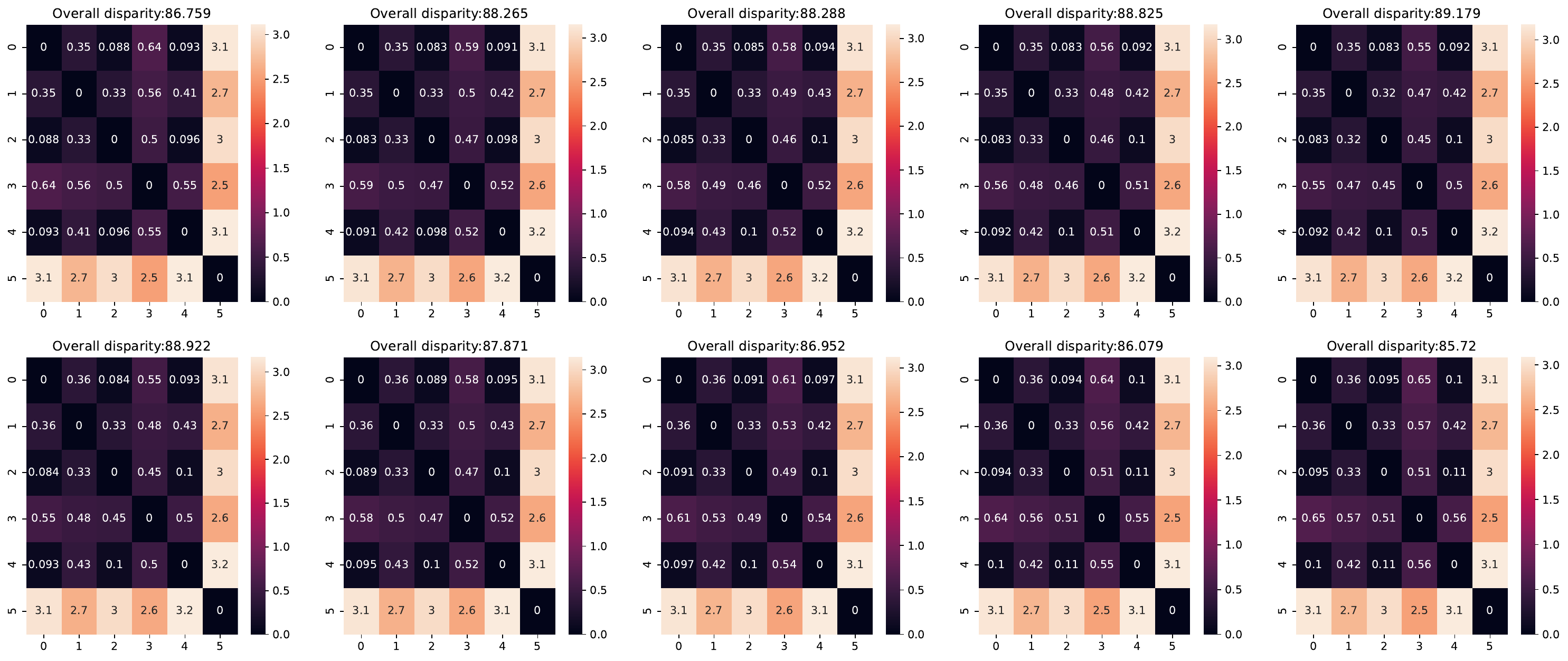}
		\caption{Scale Top feature in COIL-20.}
	\end{subfigure}
	\caption{Distance matrices for redundant features. We select the top 5 relevant features at the beginning, then sequentially copy part of the selected features. (a) shows the distance matrices with the increase of redundant features, (b) shows the distance matrices after dividing the mean of each matrix.}
	\label{dist_mat_copy_top_activity}
\end{figure}

\subsection{Experiment 2: when features are redundant}

Studying the similarity or redundancy between features is a fundamental problem in statistics and machine learning. Our methods provide an alternative to the typical correlation-based measures in that we explore redundancy in features through the similarity between their distance matrices. We consider scenarios that compare the similarity between feature sets and the influence of linearly transformed duplicate features on the distance matrices of selected features to validate our methods in detecting redundancy in features.

\begin{figure}[!ht]
	\centering
	\begin{subfigure}[t]{0.49\textwidth}
		\includegraphics[width=\textwidth]{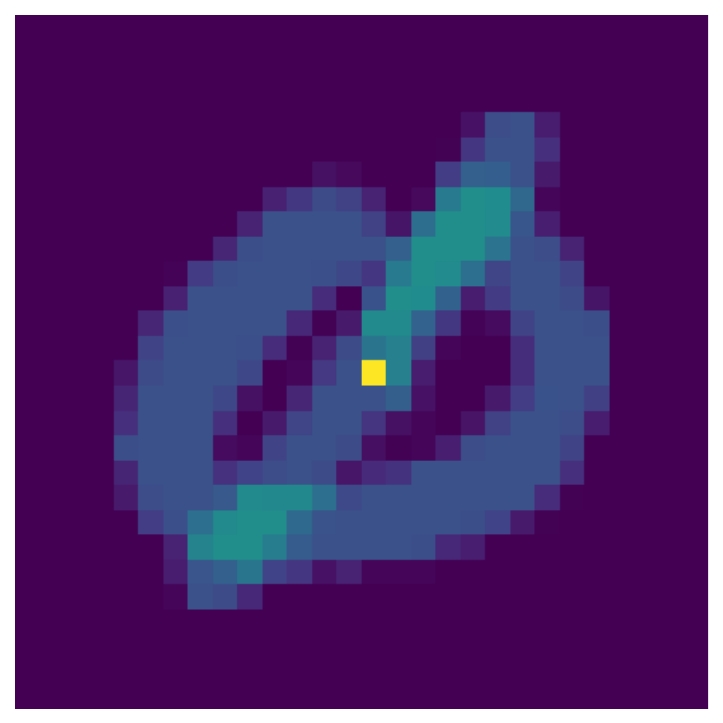}
		\caption{Images from (0,1) pair.}
	\end{subfigure}
	\begin{subfigure}[t]{0.49\textwidth}
		\includegraphics[width=\textwidth]{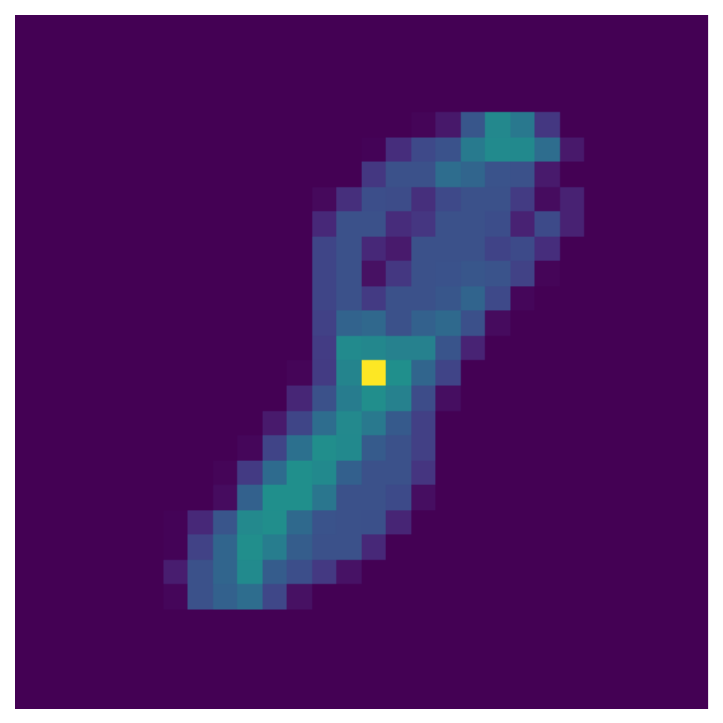}
		\caption{Images from (0,8) pair.}
	\end{subfigure}
		\begin{subfigure}[t]{0.49\textwidth}
		\includegraphics[width=\textwidth]{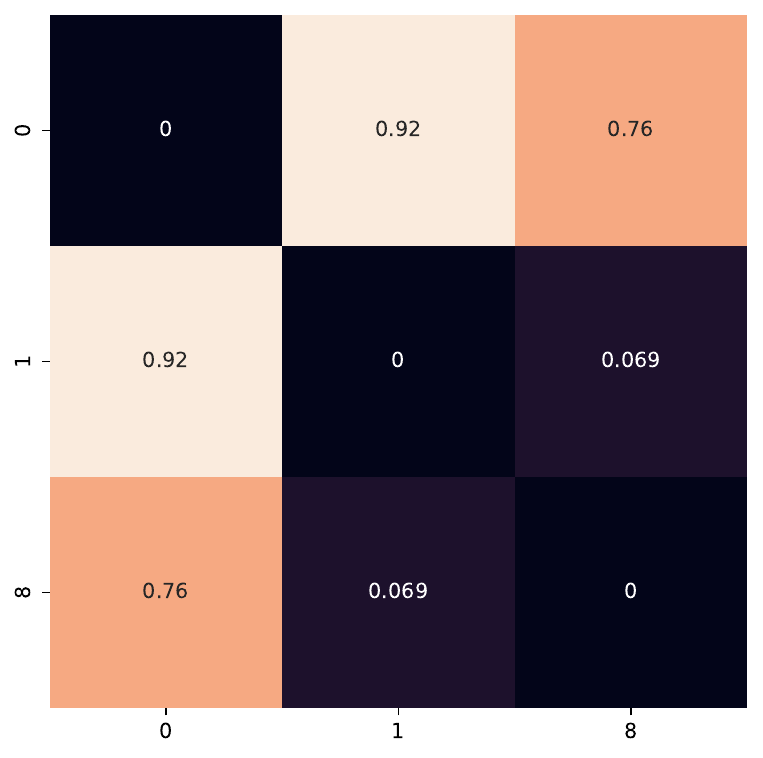}
		\caption{Distance matrix of index 406.}
	\end{subfigure}
		\begin{subfigure}[t]{0.49\textwidth}
		\includegraphics[width=\textwidth]{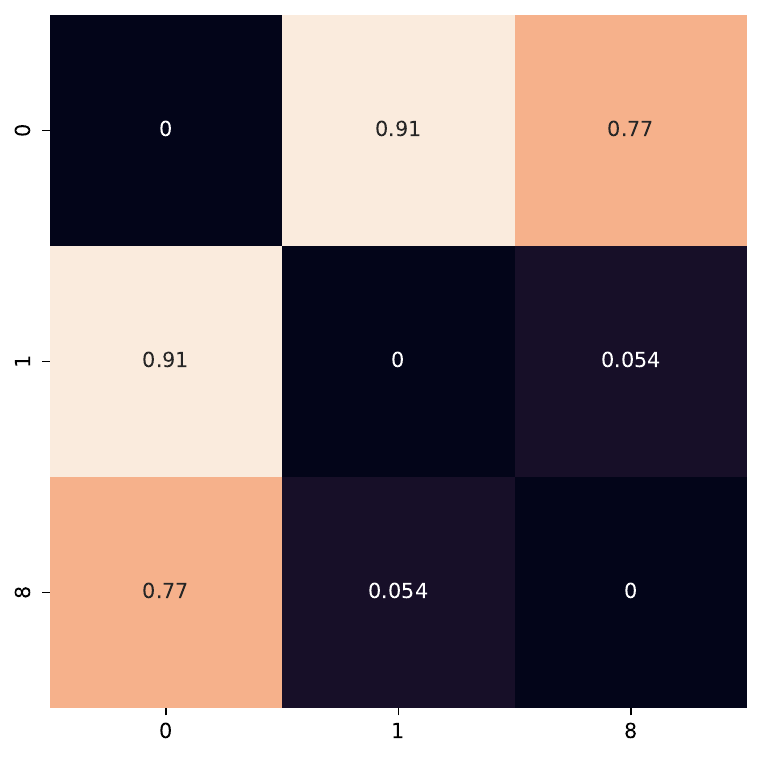}
		\caption{Distance matrix of index 434.}
	\end{subfigure}
	\caption{Demo of exploring relevance and redundancy by distance matrices.}
	\label{demo_mnist_dist_mat_central_1fs}
\end{figure}

As shown in Fig. \ref{demo_mnist_dist_mat_central_1fs}, the pixel at the center of the images could be used to distinguish class 0 and class 1 from the MNIST dataset, but adjacent pixels may be redundant to each other. The similarity between their distance matrices provides a quantitative measure of redundancy that could be used in further analysis.

We also validate the effectiveness of redundancy detection by data matrices using duplicate features. We sequentially copy a portion of the selected feature set. The results in Fig. \ref{dist_mat_copy_top_activity} show that duplicate features from the Activity dataset change the disparity between classes, but it almost disappears when we scale the distance matrix by the mean of its values. This means that the distance matrix, which consists of the distances between class conditional distributions, captures the disparity of new features, and it also shows that distance matrices could be used to measure discriminant information by means of simple scaling operation. 

\begin{figure}[!ht]
	\centering
		\includegraphics[width=\textwidth]{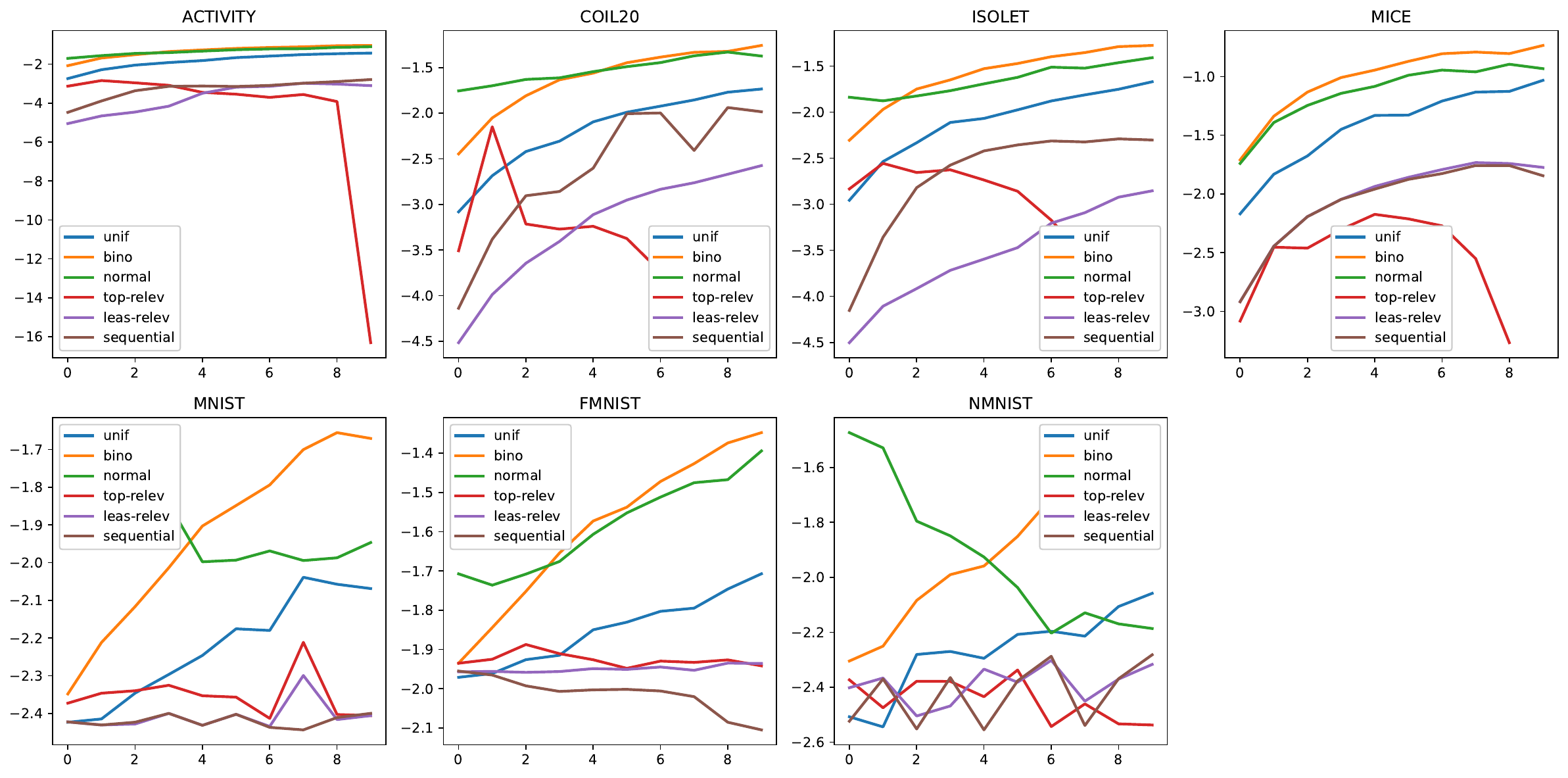}
	\caption{GWD between top-50 features with greatest disparity and selected features with the changing of size of various type of features, values are log(10) transformed for neat.}
	\label{increse_var_feature_vs_top50}
\end{figure}

We examine the distances when confronted with different types of features as another way of measuring the dissimilarity between submatrices using the Gromov-Wasserstein distance.
As shown in Fig. \ref{increse_var_feature_vs_top50}, the distances between the submatrices corresponding to the top 50 features with the greatest dissimilarity and the top relevant features tend towards zero as the number of features increases, while the distances between the submatrices corresponding to other types of features do not always tend towards zero.
For the NMNIST dataset, the dynamics are different, with the GWD decreasing with the addition of normal features, which may be due to the white Gaussian noise added in the datasets.

\begin{figure}[!ht]
	\centering
	\begin{subfigure}[b]{0.49\textwidth}
		\includegraphics[width=\textwidth]{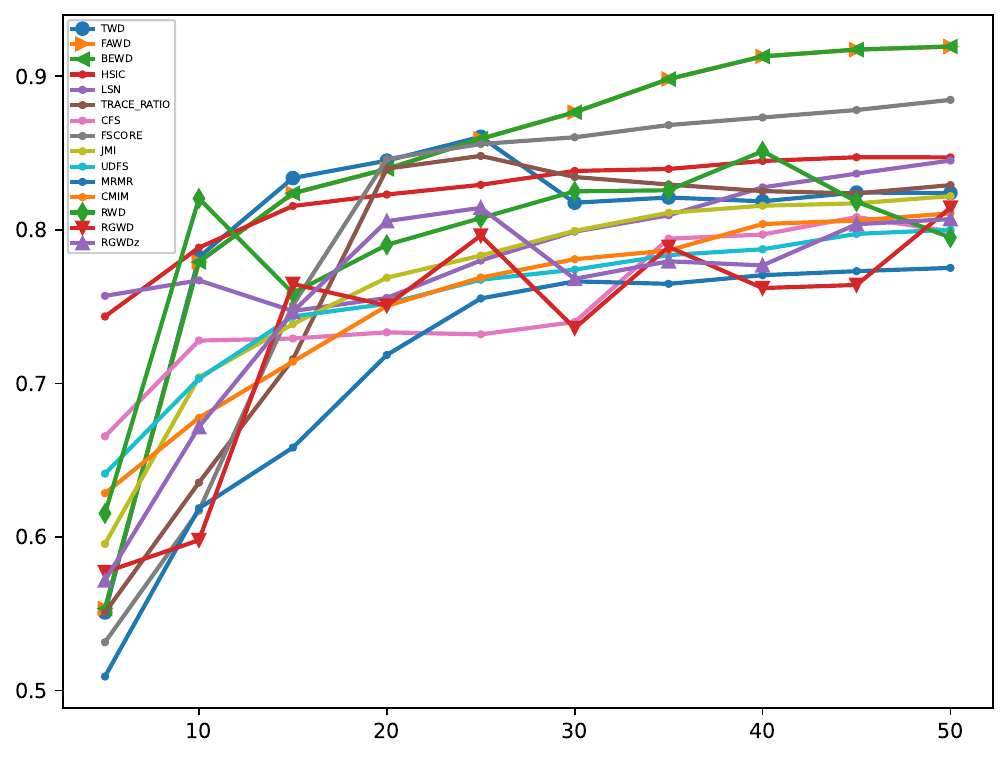}
		\caption{Activity}
		\label{Activity_xgb_acc}    
	\end{subfigure}
	\begin{subfigure}[b]{0.49\textwidth}
		\includegraphics[width=\textwidth]{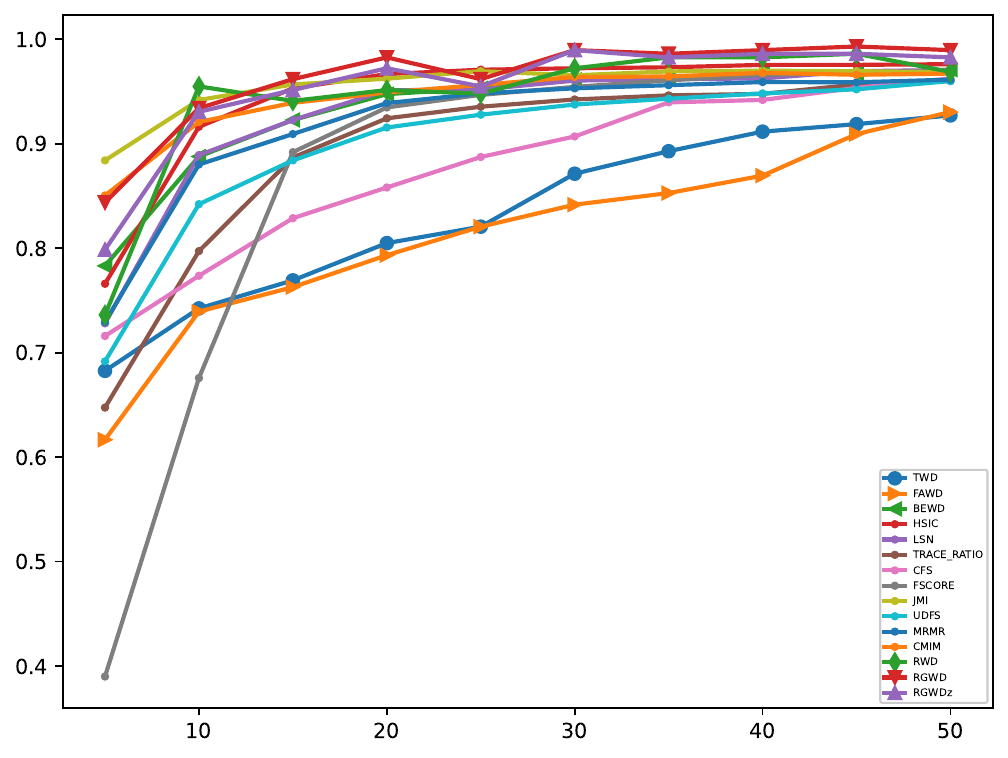}
		\caption{COIL-20}
		\label{COIL-20_xgb_acc}    
	\end{subfigure}
	\begin{subfigure}[b]{0.49\textwidth}
		\includegraphics[width=\textwidth]{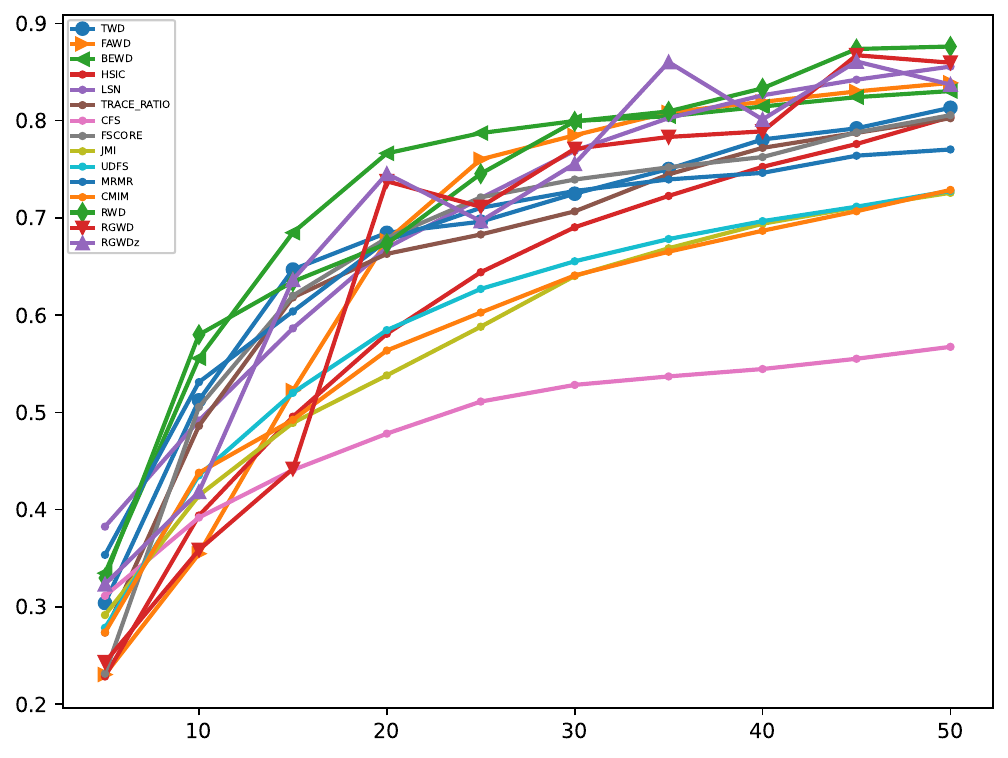}
		\caption{ISOLET}
		\label{ISOLET_xgb_acc}    
	\end{subfigure}
	\begin{subfigure}[b]{0.49\textwidth}
		\includegraphics[width=\textwidth]{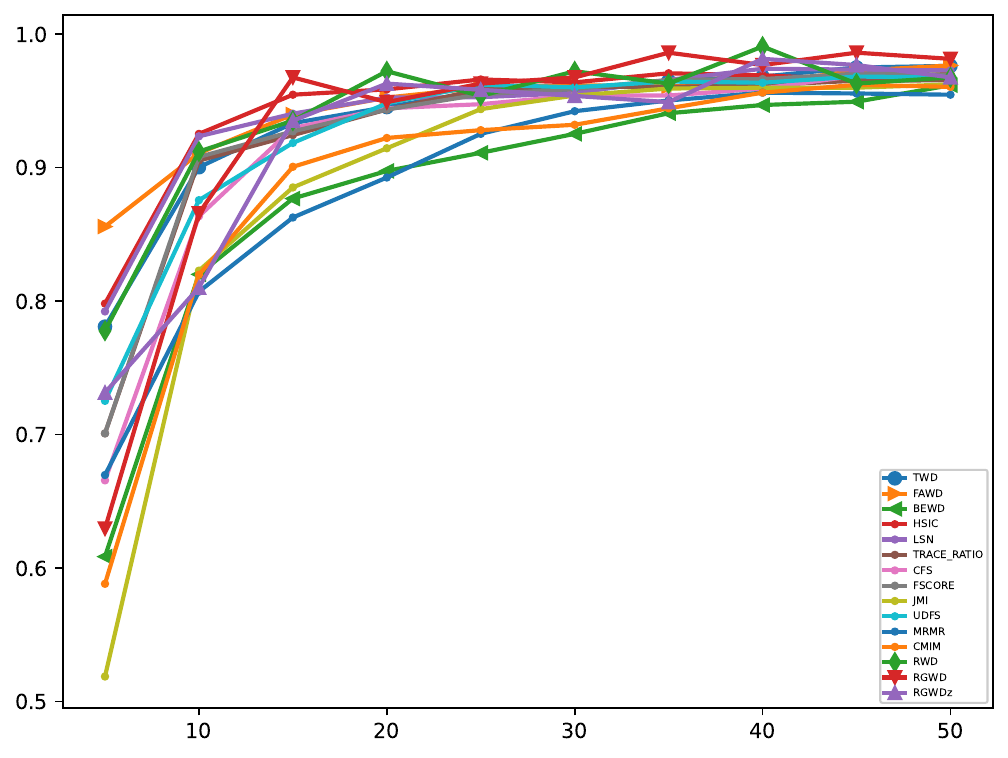}
		\caption{MICE}
		\label{MICE_xgb_acc}    
	\end{subfigure}
	\begin{subfigure}[b]{0.49\textwidth}
		\includegraphics[width=\textwidth]{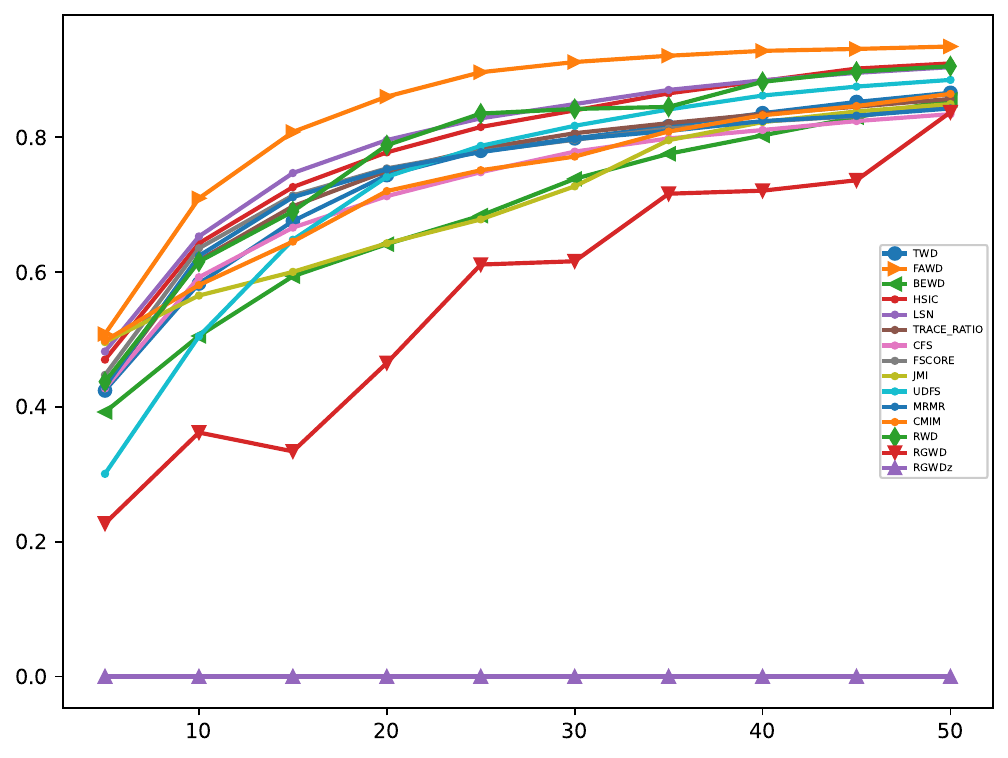}
		\caption{MNIST}
		\label{MNIST_xgb_acc}    
	\end{subfigure}
	\begin{subfigure}[b]{0.49\textwidth}
		\includegraphics[width=\textwidth]{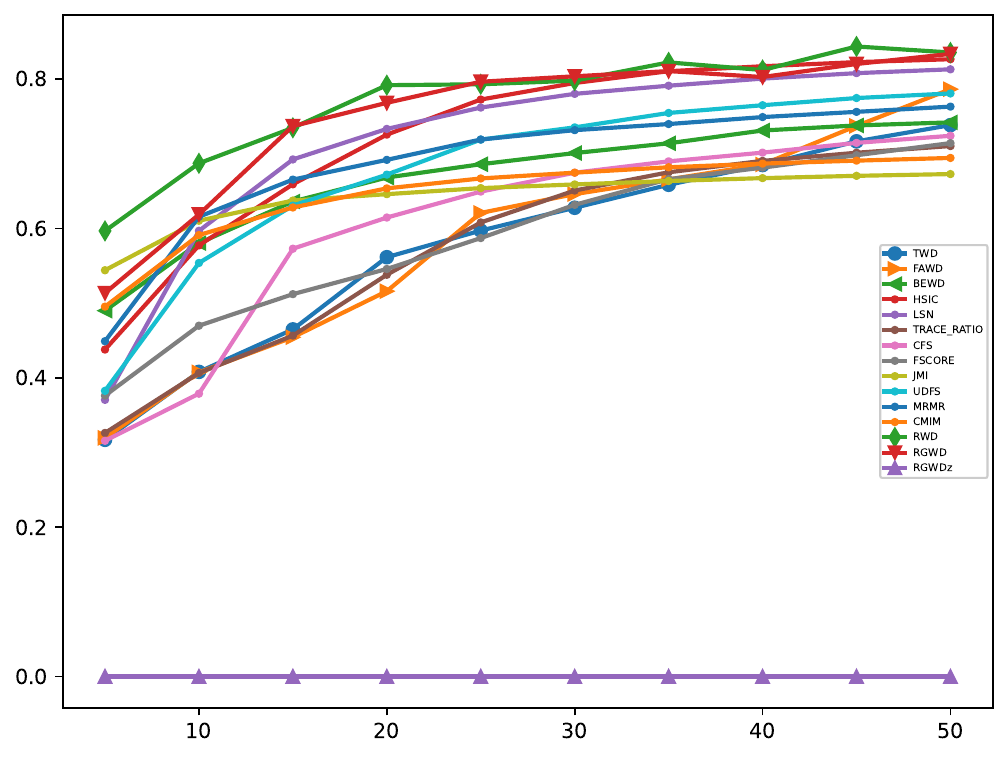}
		\caption{Fashion-MNIST}
		\label{MNIST-Fashion_xgb_acc}    
	\end{subfigure}
	\caption{Classification Accuracy.}
	\label{xgb_acc}
\end{figure}

\subsection{Experiment 3: classification accuracy}

The accuracy of classification using a set of features is a natural way of assessing the value of the features in supervised learning, and it is also commonly used to compare the performance of feature selection methods.
We use this metric to evaluate different methods by varying the number of features they select. We test the feature sets with sizes ranging from 10 to 50.

We choose XGBoost \cite{chen2016xgboost} as the downstream classifier due to its advantages in performance, interpretation, and ease of control. We implement the classifier using the xgboost Python package.
To ensure a fair comparison, we set the same hyperparameters for all feature sets, set the number of trees to 50, and set the random state of the classifiers to 0.
We use the naive random search to build a demo to show the performance of our methods to avoid the improvement from the serach strategies.

Our proposed framework, regardless of the supervised and unsupervised algorithms, performs well on all benchmark datasets from various domains, as shown in Fig. \ref{xgb_acc}. Compared to similar filter methods, our framework achieves higher accuracy performance when the number of features ranges from 5 to 50.
Compared with embedded methods such as LSN and HSIC, maximizing the disparity leads to the highest or nearly the highest accuracy on all datasets while minimizing the GWD occurs exception on the Activity dataset.
However, when eliminating the outliers in the datasets by Z-score \footnote{we set the threshold to 10, which is a mild constraint, and mark the corresponding results in the figure legend with 'z' after correction.}, the performance becomes higher, which means that eliminating outliers could be helpful for our methods.
The experimental results also show that our method is effective on datasets with small sample sizes, such as COIL-20 and MICE.

\begin{figure}[!ht]
	\centering
	\begin{subfigure}[t]{0.49\textwidth}
		\includegraphics[width=\textwidth]{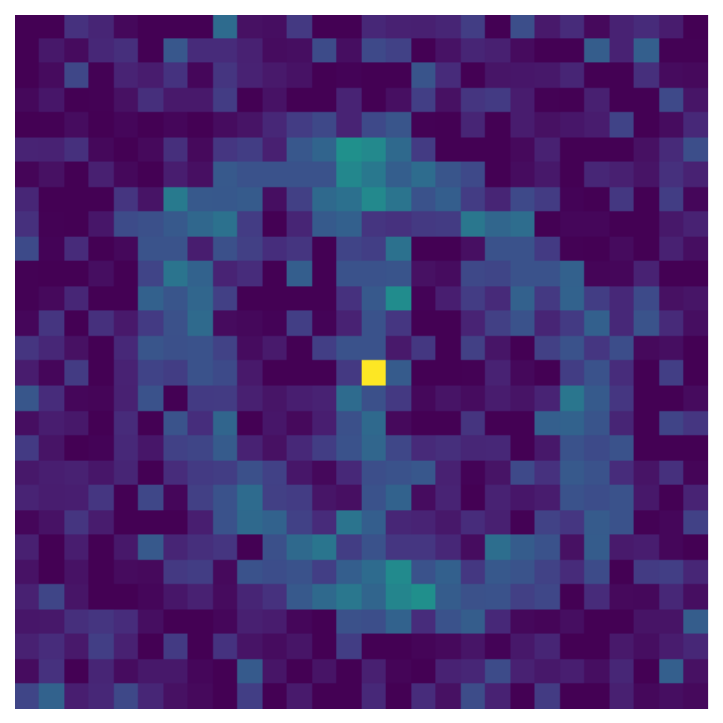}
		\caption{Images from (0,1) pair.}
	\end{subfigure}
	\begin{subfigure}[t]{0.49\textwidth}
		\includegraphics[width=\textwidth]{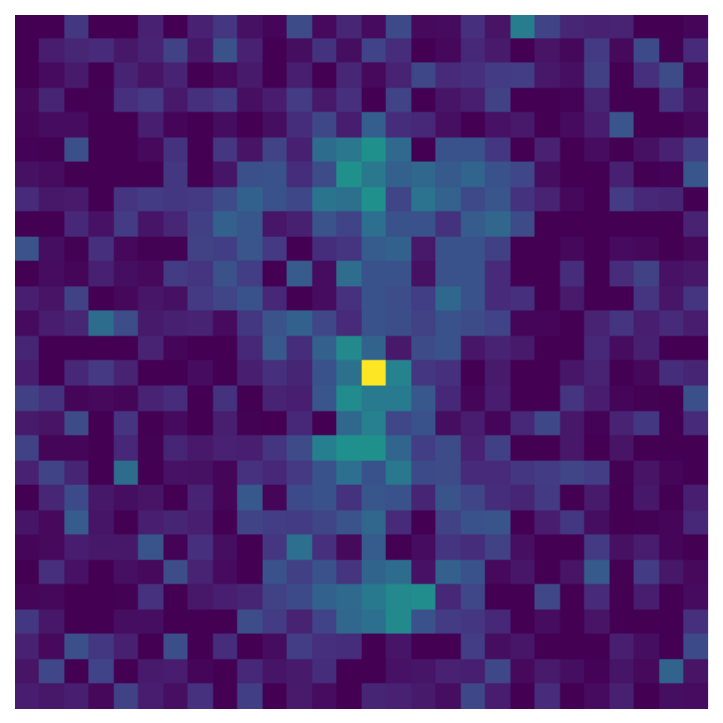}
		\caption{Images from (0,8) pair.}
	\end{subfigure}
	\begin{subfigure}[t]{0.49\textwidth}
		\includegraphics[width=\textwidth]{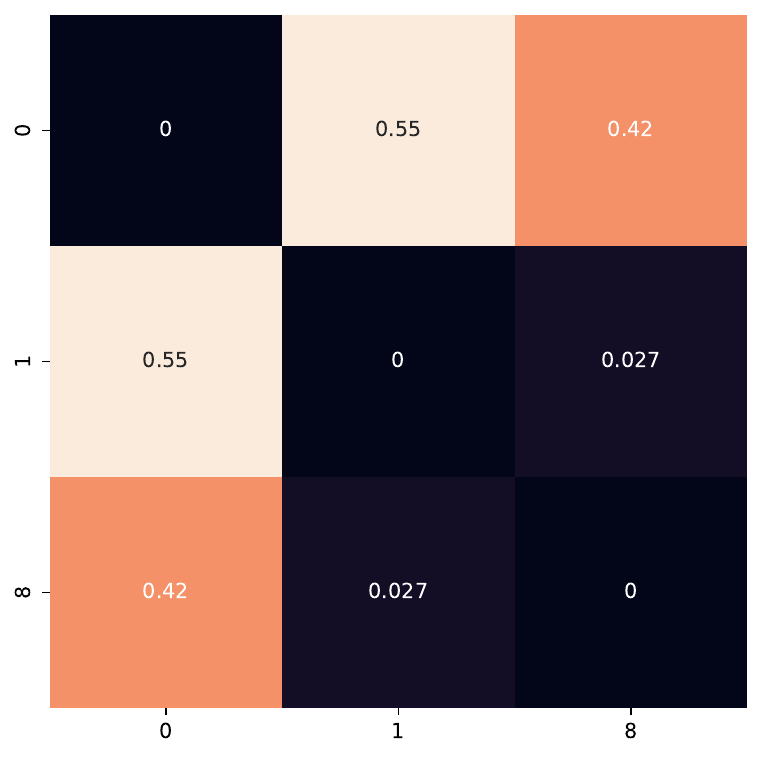}
		\caption{Distance matrix of index 406.}
	\end{subfigure}
	\begin{subfigure}[t]{0.49\textwidth}
		\includegraphics[width=\textwidth]{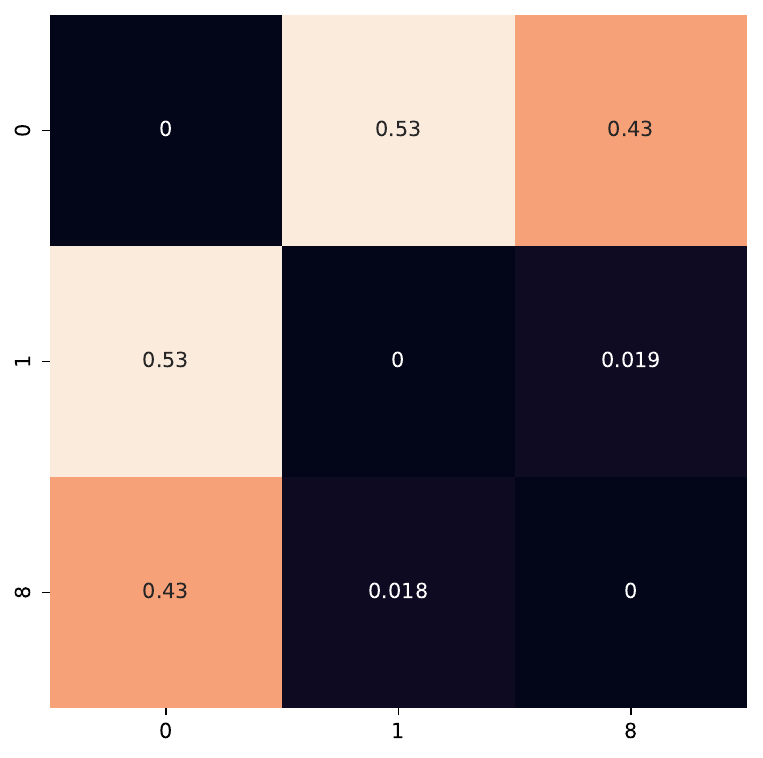}
		\caption{Distance matrix of index 434.}
	\end{subfigure}
	\caption{Demo of exploring relevance and redundancy by distance matrices.}
	\label{demo_nmnist_dist_mat_central_1fs}
\end{figure}

\subsection{Experiment 4: When dataset is noisy}

In practice, samples in datasets are often disturbed by noise; therefore, feature selection methods should still work in such settings. We implement our methods on the nMNIST-AWGN dataset, which is an MNIST-like dataset consisting of samples with the same size as the MNIST dataset. The samples are mixtures of additive white noise and images from MNIST.
Our methods still work on this dataset in both supervised and unsupervised settings.

\begin{figure}[!ht]
	\centering
	\includegraphics[width=\textwidth]{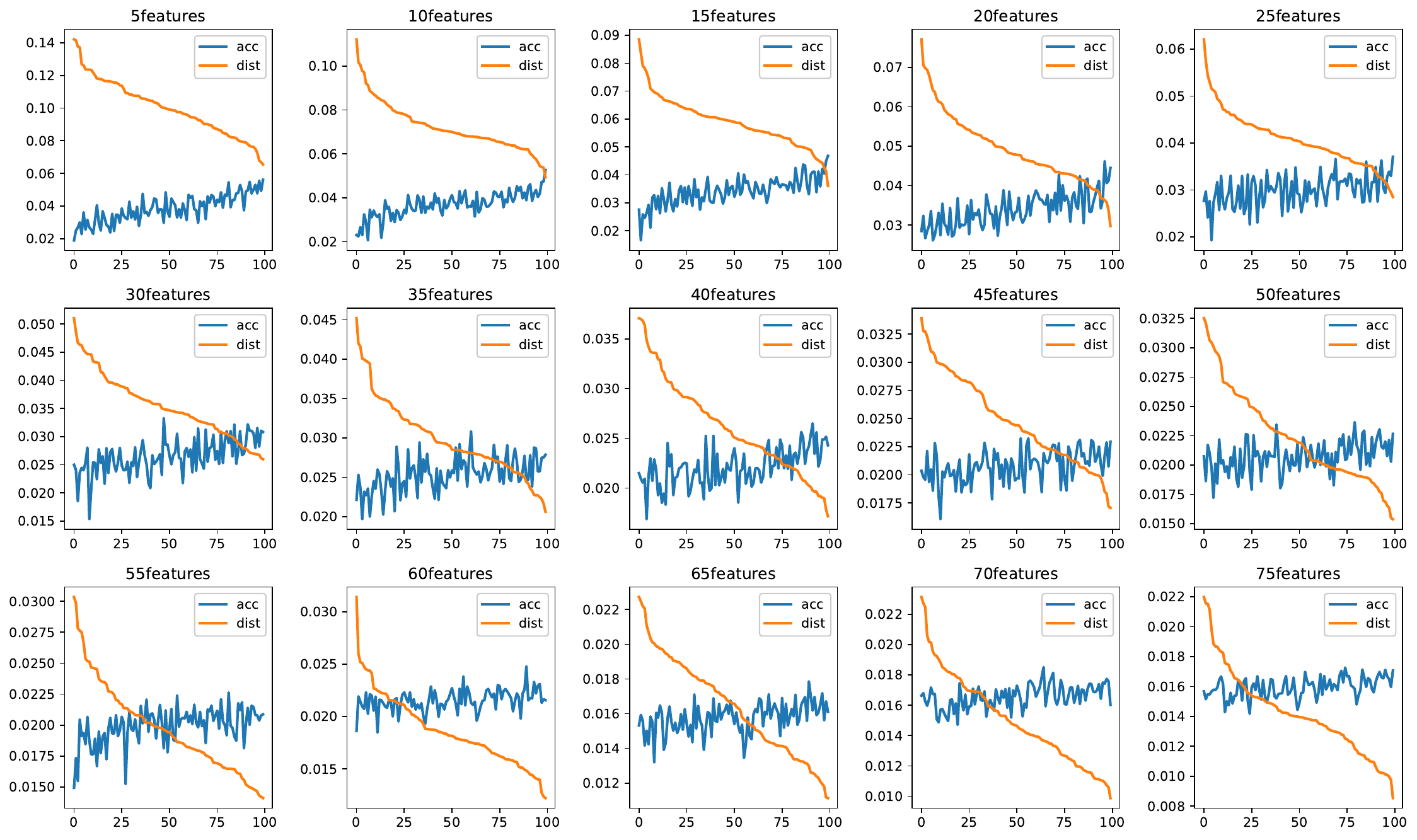}
	\caption{ GWD and accuracy corresponding to feature sets.}
	\label{gw_dist_vs_accuracy_nmnist}
\end{figure}
As previously, we explore the relevance and redundancy by means of distance matrices and GWD, we combine our criteria with random search for convenience in classification tasks, which also achieves competitive performance.
Under such a high noise setting, the distance matrices of the features could still represent the discriminative power inherent in the features, as shown in Fig. \ref{demo_nmnist_dist_mat_central_1fs}, the pixels in the center of the images show a significantly larger disparity between class 0 and class 1, while being almost irrelevant to the labels in distinguishing class 1 from class 8.
Moreover, there is a clear consistency between the reciprocal of the GWD between the submatrix and the original data matrix and the accuracy, as shown in Fig \ref{gw_dist_vs_accuracy_nmnist}, which also indicates that our criteria are effective.

\section{Conclusion}
In conclusion, this work presents a comprehensive approach to feature selection in machine learning, leveraging empirical statistics and distributions to explore feature relationships. The introduction of similarity measures for redundancy detection marks a significant advancement in the field. The proposed framework and algorithms show superior performance on benchmark datasets, providing robust solutions for measuring and reducing data redundancy. The novel redundancy measure, characterized by its noise robustness, further enhances the effectiveness of the proposed methods. This work not only contributes to the theoretical understanding of feature selection but also provides practical algorithms for both supervised and unsupervised learning settings. Future work may explore the extension of these methods to other types of data and learning scenarios.

\subsection*{Acknowledgments}

\newpage

\bibliography{feature_selection_bibt.bib}

\begin{thebibliography}{44}
\providecommand{\natexlab}[1]{#1}
\providecommand{\url}[1]{\texttt{#1}}
\expandafter\ifx\csname urlstyle\endcsname\relax
  \providecommand{\doi}[1]{doi: #1}\else
  \providecommand{\doi}{doi: \begingroup \urlstyle{rm}\Url}\fi

\bibitem[Bal{\i}n et~al.(2019)Bal{\i}n, Abid, and Zou]{balin2019concrete}
Bal{\i}n, M.~F., Abid, A., and Zou, J.
\newblock Concrete autoencoders: Differentiable feature selection and
  reconstruction.
\newblock In \emph{International Conference on Machine Learning}, pp.\
  444--453. PMLR, 2019.

\bibitem[Battiti(1994)]{battiti1994using}
Battiti, R.
\newblock Using mutual information for selecting features in supervised neural
  net learning.
\newblock \emph{IEEE Transactions on Neural Networks}, 5\penalty0 (4):\penalty0
  537--550, 1994.

\bibitem[Bellman(1957)]{bellman1957dynamic}
Bellman, R.
\newblock \emph{Dynamic Programming}.
\newblock Princeton University Press, Princeton, 1957.

\bibitem[Brown et~al.(2012)Brown, Pocock, Zhao, and
  Luj{\'a}n]{brown2012conditional}
Brown, G., Pocock, A., Zhao, M.-J., and Luj{\'a}n, M.
\newblock Conditional likelihood maximisation: A unifying framework for
  information theoretic feature selection.
\newblock \emph{Journal of Machine Learning Research}, 13\penalty0
  (2):\penalty0 27--66, 2012.

\bibitem[Cai et~al.(2018)Cai, Luo, Wang, and Yang]{cai2018feature}
Cai, J., Luo, J., Wang, S., and Yang, S.
\newblock Feature selection in machine learning: A new perspective.
\newblock \emph{Neurocomputing}, 300:\penalty0 70--79, 2018.

\bibitem[Chen \& Guestrin(2016)Chen and Guestrin]{chen2016xgboost}
Chen, T. and Guestrin, C.
\newblock Xgboost: A scalable tree boosting system.
\newblock In \emph{Proceedings of the 22nd ACM SIGKDD International Conference
  on Knowledge Discovery and Data Mining}, KDD '16, pp.\  785--794, New York,
  NY, USA, 2016. Association for Computing Machinery.
\newblock ISBN 978-1-4503-4232-2.

\bibitem[Chow \& Huang(2005)Chow and Huang]{chow2005estimating}
Chow, T. and Huang, D.
\newblock Estimating optimal feature subsets using efficient estimation of
  high-dimensional mutual information.
\newblock \emph{IEEE Transactions on Neural Networks}, 16\penalty0
  (1):\penalty0 213--224, 2005.

\bibitem[{Climente-Gonz{\'a}lez} et~al.(2019){Climente-Gonz{\'a}lez}, Azencott,
  Kaski, and Yamada]{climente-gonzalez2019block}
{Climente-Gonz{\'a}lez}, H., Azencott, C.-A., Kaski, S., and Yamada, M.
\newblock Block hsic lasso: Model-free biomarker detection for ultra-high
  dimensional data.
\newblock \emph{Bioinformatics}, 35\penalty0 (14):\penalty0 i427--i435, 2019.

\bibitem[Duda et~al.(2000)Duda, Hart, and Stork]{duda2000pattern}
Duda, R.~O., Hart, P.~E., and Stork, D.~G.
\newblock \emph{Pattern Classification}.
\newblock Wiley, New York, 2nd ed edition, 2000.
\newblock ISBN 978-0-471-05669-0.

\bibitem[Estevez et~al.(2009)Estevez, Tesmer, Perez, and
  Zurada]{estevez2009normalized}
Estevez, P., Tesmer, M., Perez, C., and Zurada, J.
\newblock Normalized mutual information feature selection.
\newblock \emph{IEEE Transactions on Neural Networks}, 20\penalty0
  (2):\penalty0 189--201, 2009.

\bibitem[Fleuret(2004)]{fleuret2004fast}
Fleuret, F.
\newblock Fast binary feature selection with conditional mutual information.
\newblock \emph{Journal of Machine Learning Research}, 5:\penalty0 1531--1555,
  2004.

\bibitem[Guyon \& Elisseeff(2003)Guyon and Elisseeff]{guyon2003introduction}
Guyon, I. and Elisseeff, A.
\newblock An introduction to variable and feature selection.
\newblock \emph{Journal of machine learning research}, 3\penalty0
  (Mar):\penalty0 1157--1182, 2003.

\bibitem[Hall(1999)]{hall1999correlationbased}
Hall, M.~A.
\newblock \emph{Correlation-Based Feature Selection for Machine Learning}.
\newblock PhD thesis, University of Waikato, 1999.

\bibitem[Hall \& Smith(1999)Hall and Smith]{hall1999feature}
Hall, M.~A. and Smith, L.~A.
\newblock Feature selection for machine learning: Comparing a correlation-based
  filter approach to the wrapper.
\newblock In \emph{FLAIRS Conference}, volume 1999, pp.\  235--239, 1999.

\bibitem[{Hanchuan Peng} et~al.(2005){Hanchuan Peng}, {Fuhui Long}, and
  Ding]{hanchuanpeng2005feature}
{Hanchuan Peng}, {Fuhui Long}, and Ding, C.
\newblock Feature selection based on mutual information criteria of
  max-dependency, max-relevance, and min-redundancy.
\newblock \emph{IEEE Transactions on Pattern Analysis and Machine
  Intelligence}, 27\penalty0 (8):\penalty0 1226--1238, 2005.

\bibitem[He et~al.(2005)He, Cai, and Niyogi]{he2005laplacian}
He, X., Cai, D., and Niyogi, P.
\newblock Laplacian score for feature selection.
\newblock In \emph{Advances in Neural Information Processing Systems},
  volume~18. MIT Press, 2005.

\bibitem[Kira \& Rendell(1992)Kira and Rendell]{kira1992practical}
Kira, K. and Rendell, L.~A.
\newblock A practical approach to feature selection.
\newblock In \emph{Proceedings of the Ninth International Workshop on Machine
  Learning}, ML92, pp.\  249--256, San Francisco, CA, USA, 1992. Morgan
  Kaufmann Publishers Inc.

\bibitem[Kohavi \& John(1997)Kohavi and John]{kohavi1997wrappers}
Kohavi, R. and John, G.~H.
\newblock Wrappers for feature subset selection.
\newblock \emph{Artificial Intelligence}, 97\penalty0 (1-2):\penalty0 273--324,
  1997.

\bibitem[Lemhadri et~al.(2021)Lemhadri, Ruan, Abraham, and
  Tibshirani]{lemhadri2021lassonet}
Lemhadri, I., Ruan, F., Abraham, L., and Tibshirani, R.
\newblock Lassonet: A neural network with feature sparsity.
\newblock \emph{Journal of Machine Learning Research}, 22\penalty0 (127), 2021.

\bibitem[Lewis(1992)]{lewis1992feature}
Lewis, D.~D.
\newblock Feature selection and feature extraction for text categorization.
\newblock In \emph{Proceedings of the Workshop on Speech and Natural Language},
  HLT '91, pp.\  212--217, USA, 1992. Association for Computational
  Linguistics.
\newblock ISBN 978-1-55860-272-4.

\bibitem[Li et~al.(2018)Li, Cheng, Wang, Morstatter, Trevino, Tang, and
  Liu]{li2018feature}
Li, J., Cheng, K., Wang, S., et~al.
\newblock Feature selection: A data perspective.
\newblock \emph{ACM Computing Surveys}, 50\penalty0 (6):\penalty0 1--45, 2018.

\bibitem[Li et~al.(2012)Li, Yang, Liu, Zhou, and Lu]{li2012unsupervised}
Li, Z., Yang, Y., Liu, J., Zhou, X., and Lu, H.
\newblock Unsupervised feature selection using nonnegative spectral analysis.
\newblock \emph{Proceedings of the AAAI Conference on Artificial Intelligence},
  26\penalty0 (1):\penalty0 1026--1032, 2012.

\bibitem[M{\'e}moli(2011)]{memoli2011gromov}
M{\'e}moli, F.
\newblock Gromov{\textendash}wasserstein distances and the metric approach to
  object matching.
\newblock \emph{Foundations of Computational Mathematics}, 11\penalty0
  (4):\penalty0 417--487, 2011.

\bibitem[Meyer \& Bontempi(2006)Meyer and Bontempi]{meyer2006use}
Meyer, P.~E. and Bontempi, G.
\newblock On the use of variable complementarity for feature selection in
  cancer classification.
\newblock In Rothlauf, F., Branke, J., Cagnoni, S., et~al. (eds.),
  \emph{Applications of Evolutionary Computing}, Lecture Notes in Computer
  Science, pp.\  91--102, Berlin, Heidelberg, 2006. Springer.
\newblock ISBN 978-3-540-33238-1.

\bibitem[Meyer et~al.(2008)Meyer, Schretter, and
  Bontempi]{meyer2008informationtheoretic}
Meyer, P.~E., Schretter, C., and Bontempi, G.
\newblock Information-theoretic feature selection in microarray data using
  variable complementarity.
\newblock \emph{IEEE Journal of Selected Topics in Signal Processing},
  2\penalty0 (3):\penalty0 261--274, 2008.

\bibitem[Nguyen et~al.(2009)Nguyen, Wainwright, and
  Jordan]{nguyen2009surrogate}
Nguyen, X., Wainwright, M.~J., and Jordan, M.~I.
\newblock On surrogate loss functions and f-divergences.
\newblock \emph{The Annals of Statistics}, 37\penalty0 (2):\penalty0 876--904,
  2009.

\bibitem[Nie et~al.(2008)Nie, Xiang, Jia, Zhang, and Yan]{nie2008trace}
Nie, F., Xiang, S., Jia, Y., Zhang, C., and Yan, S.
\newblock Trace ratio criterion for feature selection.
\newblock In \emph{AAAI}, volume~2, pp.\  671--676, 2008.

\bibitem[Novovicova et~al.(1996)Novovicova, Pudil, and
  Kittler]{novovicova1996divergence}
Novovicova, J., Pudil, P., and Kittler, J.
\newblock Divergence based feature selection for multimodal class densities.
\newblock \emph{IEEE Transactions on Pattern Analysis and Machine
  Intelligence}, 18\penalty0 (2):\penalty0 218--223, 1996.

\bibitem[P{\'o}czos et~al.(2012)P{\'o}czos, Ghahramani, and
  Schneider]{poczos2012copulabased}
P{\'o}czos, B., Ghahramani, Z., and Schneider, J.
\newblock Copula-based kernel dependency measures.
\newblock In \emph{Proceedings of the 29th International Coference on
  International Conference on Machine Learning}, ICML'12, pp.\  1635--1642,
  Madison, WI, USA, 2012. Omnipress.
\newblock ISBN 978-1-4503-1285-1.

\bibitem[Song et~al.(2012)Song, Smola, Gretton, Bedo, and
  Borgwardt]{song2012feature}
Song, L., Smola, A., Gretton, A., Bedo, J., and Borgwardt, K.
\newblock Feature selection via dependence maximization.
\newblock \emph{Journal of Machine Learning Research}, 13\penalty0 (5), 2012.

\bibitem[Sriperumbudur et~al.(2009)Sriperumbudur, Fukumizu, Gretton,
  Sch{\"o}lkopf, and Lanckriet]{sriperumbudur2009integral}
Sriperumbudur, B.~K., Fukumizu, K., Gretton, A., Sch{\"o}lkopf, B., and
  Lanckriet, G. R.~G.
\newblock On integral probability metrics, {\textbackslash}phi-divergences and
  binary classification, 2009.

\bibitem[Tuv et~al.(2009)Tuv, Borisov, Runger, and Torkkola]{tuv2009feature}
Tuv, E., Borisov, A., Runger, G., and Torkkola, K.
\newblock Feature selection with ensembles, artificial variables, and
  redundancy elimination.
\newblock \emph{Journal of Machine Learning Research}, 10\penalty0
  (45):\penalty0 1341--1366, 2009.

\bibitem[Wollstadt et~al.(2023)Wollstadt, Schmitt, and
  Wibral]{wollstadt2023rigorous}
Wollstadt, P., Schmitt, S., and Wibral, M.
\newblock A rigorous information-theoretic definition of redundancy and
  relevancy in feature selection based on (partial) information decomposition.
\newblock \emph{Journal of Machine Learning Research}, 24\penalty0
  (131):\penalty0 1--44, 2023.

\bibitem[Wu \& Cheng(2021)Wu and Cheng]{wu2021fractal}
Wu, X. and Cheng, Q.
\newblock Fractal autoencoders for feature selection.
\newblock \emph{Proceedings of the AAAI Conference on Artificial Intelligence},
  35\penalty0 (12):\penalty0 10370--10378, 2021.

\bibitem[Xu et~al.(2017)Xu, Tang, He, and Man]{xu2017semisupervised}
Xu, J., Tang, B., He, H., and Man, H.
\newblock Semisupervised feature selection based on relevance and redundancy
  criteria.
\newblock \emph{IEEE Transactions on Neural Networks and Learning Systems},
  28\penalty0 (9):\penalty0 1974--1984, 2017.

\bibitem[Xu et~al.(2022)Xu, Wu, Wei, Zhong, and Nie]{xu2022general}
Xu, X., Wu, X., Wei, F., Zhong, W., and Nie, F.
\newblock A general framework for feature selection under orthogonal regression
  with global redundancy minimization.
\newblock \emph{IEEE Transactions on Knowledge and Data Engineering},
  34\penalty0 (11):\penalty0 5056--5069, 2022.

\bibitem[Yamada et~al.(2014)Yamada, Jitkrittum, Sigal, Xing, and
  Sugiyama]{yamada2014highdimensional}
Yamada, M., Jitkrittum, W., Sigal, L., Xing, E.~P., and Sugiyama, M.
\newblock High-dimensional feature selection by feature-wise kernelized lasso.
\newblock \emph{Neural Computation}, 26\penalty0 (1):\penalty0 185--207, 2014.

\bibitem[Yang \& Moody(1999)Yang and Moody]{yang1999data}
Yang, H. and Moody, J.
\newblock Data visualization and feature selection: New algorithms for
  nongaussian data.
\newblock In \emph{Advances in Neural Information Processing Systems},
  volume~12. MIT Press, 1999.

\bibitem[Yang et~al.(2011)Yang, Shen, Ma, Huang, and Zhou]{yang2011l2}
Yang, Y., Shen, H.~T., Ma, Z., Huang, Z., and Zhou, X.
\newblock L2, 1-norm regularized discriminative feature selection for
  unsupervised.
\newblock In \emph{Twenty-Second International Joint Conference on Artificial
  Intelligence}, 2011.

\bibitem[Yang \& Pedersen(1997)Yang and Pedersen]{yang1997comparative}
Yang, Y. and Pedersen, J.~O.
\newblock A comparative study on feature selection in text categorization.
\newblock In \emph{Proceedings of the Fourteenth International Conference on
  Machine Learning}, ICML '97, pp.\  412--420, San Francisco, CA, USA, 1997.
  Morgan Kaufmann Publishers Inc.
\newblock ISBN 978-1-55860-486-5.

\bibitem[Yu \& Liu(2003)Yu and Liu]{yu2003feature}
Yu, L. and Liu, H.
\newblock Feature selection for high-dimensional data: A fast correlation-based
  filter solution.
\newblock In \emph{Proceedings of the Twentieth International Conference on
  International Conference on Machine Learning}, ICML'03, pp.\  856--863,
  Washington, DC, USA, 2003. AAAI Press.
\newblock ISBN 978-1-57735-189-4.

\bibitem[Yu \& Liu(2004)Yu and Liu]{yu2004efficient}
Yu, L. and Liu, H.
\newblock Efficient feature selection via analysis of relevance and redundancy.
\newblock \emph{Journal of Machine Learning Research}, 5\penalty0
  (Oct):\penalty0 1205--1224, 2004.

\bibitem[Zhao \& Liu(2007)Zhao and Liu]{zhao2007spectral}
Zhao, Z. and Liu, H.
\newblock Spectral feature selection for supervised and unsupervised learning.
\newblock In \emph{Proceedings of the 24th International Conference on Machine
  Learning}, ICML '07, pp.\  1151--1157, New York, NY, USA, 2007. Association
  for Computing Machinery.
\newblock ISBN 978-1-59593-793-3.

\bibitem[Zhao et~al.(2013)Zhao, Wang, Liu, and Ye]{zhao2013similarity}
Zhao, Z., Wang, L., Liu, H., and Ye, J.
\newblock On similarity preserving feature selection.
\newblock \emph{IEEE Transactions on Knowledge and Data Engineering},
  25\penalty0 (3):\penalty0 619--632, 2013.

\end{thebibliography}

\bibliographystyle{nips-no-url}
\newpage
\end{spacing}

\end{document}